\documentclass[a4paper,fleqn]{cas-dc}
\usepackage[numbers]{natbib}
\usepackage{amsmath,amsfonts,amssymb}
\usepackage{bbding,pifont}
\usepackage{float}
\usepackage{stfloats}
\usepackage{subcaption}
\usepackage{multirow}
\usepackage{pifont}% http://ctan.org/pkg/pifont
\usepackage{booktabs}
\usepackage[ruled,linesnumbered]{algorithm2e}
\usepackage{xcolor}
\usepackage{lscape}
\def\tsc#1{\csdef{#1}{\textsc{\lowercase{#1}}\xspace}}
\tsc{WGM}
\tsc{QE}
\tsc{EP}
\tsc{PMS}
\tsc{BEC}
\tsc{DE}
\begin{document}

\let\WriteBookmarks\relax
\def\floatpagepagefraction{1}
\def\textpagefraction{.001}

\shorttitle{Feature Reweighting for EEG-based Motor Imagery Classification}

\shortauthors{Taveena Lotey et~al.}

\title [mode = title]{Feature Reweighting for EEG-based Motor Imagery Classification}      

% \tnotemark[1,2]


% \tnotetext[1]{This document is the results of the research
%    project funded by the National Science Foundation.}

% \tnotetext[2]{The second title footnote which is a longer text matter
%    to fill through the whole text width and overflow into
%    another line in the footnotes area of the first page.}

\author[1]{Taveena Lotey}[
% type=editor,
                        % auid=000,bioid=1,
                        % prefix=Sir,
                        % role=Researcher,
                        % orcid=0000-0001-7511-2910
                        ]

% \cormark[1]


\ead{taveena@cs.iitr.ac.in}

% \ead[url]{www.cvr.cc, cvr@sayahna.org}

\credit{Writing – original draft, Conceptualization, Methodology, Formal analysis, Visualization}

\affiliation[1]{organization={Indian Institute of Technology Roorkee},
    % addressline={Radarweg 29}, 
    % city={Roorkee},
    % citysep={}, % Uncomment if no comma needed between city and postcode
    % postcode={247667}, 
    % state={Uttarakhand},
    country={India}}

\author[1]{Prateek Keserwani}[
                                % style=chinese
                                ]
\ead{pkeserwani@cs.iitr.ac.in}
\credit{Writing - Original draft preparation, Methodology, Investigation, Validation}
\author[2]{Debi Prosad Dogra}[%
   % role=Co-ordinator,
   % suffix=Jr,
   ]
\ead{dpdogra@iitbbs.ac.in}
% \ead[URL]{www.sayahna.org}

\credit{Writing – review & editing, Validation}

\affiliation[2]{organization={Indian Institute of Technology Bhubaneswar},
    % city={Bhubaneswar},
    % citysep={}, % Uncomment if no comma needed between city and postcode
    % postcode={752050}, 
    % state={Odisha},
    country={India}}

\author%
[1]
{Partha Pratim Roy}
\cormark[1]
% \fnmark[1,3]
\ead{partha@cs.iitr.ac.in}

\credit{Supervision, Writing – review & editing, Validation}
% \ead[URL]{www.stmdocs.in}

% \affiliation[3]{organization={STM Document Engineering Pvt Ltd.},
%     addressline={Mepukada}, 
%     city={Malayinkil},
%     % citysep={}, % Uncomment if no comma needed between city and postcode
%     postcode={695571}, 
%     state={Trivandrum},
%     country={India}}

\cortext[cor1]{Corresponding author}
\begin{abstract}
Classifying motor imagery (MI) using non-invasive electroencephalographic (EEG) signals is crucial for predicting the intention of limb movements. \textcolor{black}{Convolutional neural networks (CNNs) have been widely adopted for MI-EEG classification, but challenges such as low signal-to-noise ratio, non-stationarity, and non-linearity of EEG signals, along with the presence of irrelevant information in feature maps, can degrade performance.} This work proposes a novel feature reweighting approach to address these issues by introducing a feature reweighting module that suppresses irrelevant temporal and channel features. \textcolor{black}{The module generates relevance scores to reweight feature maps, thereby reducing the influence of noise and irrelevant data.} Experimental results demonstrate significant improvements in MI-EEG classification on the Physionet motor imagery and BCI Competition IV-2a datasets, achieving performance gains of 9.34\% and 3.82\%, respectively, over state-of-the-art CNN-based methods. \textcolor{black}{Furthermore, the proposed method showed competitive or superior performance on both speech imagery and motor movement tasks, highlighting its generalizability and robustness.}

\end{abstract}

% \includegraphics{images/proposed/grabs.pdf}

% \begin{highlights}
% \item Proposed a novel feature reweighting approach to enhance the classification of motor imagery EEG signals by suppressing irrelevant information. 

% % \item \textcolor{black}{The proposed convolutional neural network-based approach learns the relevance score and then reweights the features to suppress the irrelevant features and enhance the relevant ones. }

% \item The impact of each module in the proposed network is computed, using two publicly available MI datasets.

% \item \textcolor{black}{The ablation study suggests that reweighting features across both temporal and channel dimensions improves classification performance.}

% \item \textcolor{black}{The generalizability of the proposed method was validated on additional task datasets, including speech imagery and motor movement tasks.}
% \end{highlights}


% \end{frontmatter}


\maketitle

\section{Introduction}
\label{sec:introduction}
Brain-Computer Interface (BCI) has opened a new dimension for enclosing the proximity between machines and humans. BCI technologies help to establish communication between external devices and the human brain. Electroencephalography (EEG) is a highly preferred type of BCI as it is a high temporal resolution technology with non-invasiveness, cost efficiency, and easy portability characteristics \cite{wu2020transfer,cincotti2008non}. 
EEG-based BCI systems include different paradigms, such as P300, steady-state visual evoked potentials (SSVEPs), and motor imagery (MI). Among these, MI has attracted the interest of researchers to a large extent.
MI is defined as the spontaneous motor intention/imagination of limb movement (such as hands and legs) without any external stimuli. MI classification corresponds to the identification of the type of spontaneous motor intention/imagery. A wide range of applications of MI includes rehabilitation after motor disabilities \cite{pichiorri2015brain,shih2012brain}, smart home applications \cite{kosmyna2016feasibility}, robotic arm movements \cite{liu2018motor, onose2012feasibility}, complex cursor control \cite{sitaram2007temporal} and BCI-controlled games \cite{dai2020hs,gong2021deep}. 

% , sharma2022motor , al2021deep , kim2020bci
The significant challenges in MI-EEG signal classification are non-linearity, non-stationarity, high complexity, and low signal-to-noise ratio (SNR) of EEG signals \cite{jiang2019removal}. 
The other challenges are physiological artifacts introduced during EEG acquisition, such as eye and muscle movement. Moreover, psychological factors such as depression, anxiety, current mood, motivation, and attention may also affect BCI performance \cite{jiang2019removal,nijboer2008auditory,nijboer2010influence,saha2017evidence}.
Although EEG signals have a high temporal resolution, various artifacts introduce noise. These challenges complicate EEG signal decoding for MI classification, and thus, extracting accurate features from the signals becomes a highly intricate task. 

In the past, many works were performed to classify complex MI-EEG signals. These methods employed either multistage traditional machine learning methodology \cite{Blankertz2007non,Ang2012filter,yang2021motor} or end-to-end deep learning techniques \cite{lawhern2018eegnet, schirrmeister2017deep}. The prominent convolutional neural network (CNN) based deep learning methods include EEGNet \cite{lawhern2018eegnet}, ShallowConvNet \cite{schirrmeister2017deep} and DeepConvNet \cite{schirrmeister2017deep}. These methods try to handle the low SNR of EEG signals to an extent, yet the classification performance could be better. Thus, an efficient noise reduction mechanism is required to improve MI-EEG classification.

\textcolor{black}{The CNN-based deep learning models \cite{lawhern2018eegnet,schirrmeister2017deep} are a stack of various convolution operations. These methods combine the activation of all EEG electrodes into a set of features by applying convolution operation. This collection of features along the temporal sequence of the EEG signal is known as feature maps. Thus, MI-EEG signals fed to a CNN-based model get transformed into temporal feature maps. These feature maps are used to classify the MI-EEG signals. However, if the CNN is applied to noisy data, it produces noisy temporal feature maps containing irrelevant information. Additionally, convolution operation in a CNN-based model computes a feature map based on all previous feature maps, i.e., a single feature map values are computed by convolving a kernel/filter on all channels of the previous feature maps \cite{lecun2015deep}. Hence, the noise accumulates from channels of one feature map to another while applying convolution operations. }

Hence, a noise reduction mechanism is required to emphasize the relevant temporal and channel information of the feature maps of CNN to obtain an efficient EEG signal decoding for the MI task. \textit{Feature reweighting} has been used in the computer vision domain to extract useful information and suppress irrelevant information \cite{kang2019few,zhao20193d}. Inspired by \cite{kang2019few,zhao20193d}, a novel feature reweighting approach for MI-EEG signals classification is introduced in this work. \textcolor{black}{This approach automatically computes the relevance score of the features, and then reweights the feature maps with the computed relevance score. Reweighting the features with relevance score suppresses the less informative features and retains the more informative features.}

The proposed method computes a score of the relevance of features for improved EEG data classification. The score is computed over two aspects of the data, i.e., \textcolor{black}{temporal and channel}. The proposed approach computes the \textcolor{black}{relevance scores of temporal and channel features through two sub-modules: the temporal feature score (TFS) module and the channel feature score (CFS) module.} As mentioned in \cite{li2021temporal}, the direct combination of features causes information redundancy, and \textcolor{black}{information redundancy leads to overfitting}. Thus, a score fusion module is introduced in the proposed method to combine the temporal and channel feature scores efficiently. The fused scores are then used to reweight the feature maps to suppress irrelevant information. \textcolor{black}{The ablation study conducted in this work confirms that features reweighted with the relevance score lead to better feature extraction that results in improved classification.}

The proposed method outperforms the compared methods on two motor imagery datasets by significant margins. It indicates that the proposed method can extract relevant features and suppress irrelevant features for improved MI-EEG classification. \textcolor{black}{Also, the proposed method have performed exceptionally well for speech imagery task and competitively for motor movement task. It shows the generalizability and robustness of the proposed method as it performed well in tasks other than MI. }

The primary contributions of this work are as follows:
\begin{enumerate}
 
\item 
We propose a novel feature reweighting approach to enhance the classification of MI-EEG signals by suppressing irrelevant information in the EEG data.

\item 
\textcolor{black}{This work introduces two relevance score modules$-$ Temporal Feature Score (TFS) and Channel Feature Score (CFS)$-$which automatically learn the relevance of features across temporal and channel dimensions. A Score Fusion (SF) module efficiently combines these scores to generate reweighting scores for both temporal and channel features.}

\item \textcolor{black}{We conducted comprehensive ablation experiments to evaluate the impact of each module of the proposed network on EEG signal classification performance through two famous public MI datasets: BCI Competition IV-2a and the Physionet EEG motor movement/imagery database.}

\item \textcolor{black}{The proposed method was tested on both speech imagery and motor movement tasks, where it showed competitive or superior performance, demonstrating its generalizability and robustness.}

\end{enumerate}

The rest of the paper is organized into the following sections. Section \ref{sec:related_work} covers the description of existing works related to the proposed method. In Section \ref{sec:proposed_method}, \textcolor{black}{details of the proposed method and network have been discussed.} The experimental details and the performance analysis are provided in Section \ref{sec:experiments}. \textcolor{black}{The ablation study and Grad-CAM analysis of the proposed method is discussed} in Section \ref{sec:discussion}. The conclusions are presented in Section \ref{sec:conclusion}.

\section{Related Work}
\label{sec:related_work}
For classification of MI EEG signals, two types of methods have been majorly used in the literature in recent years. These works are either machine learning or deep learning based classification methods. Therefore, this section discusses the works related to the machine learning and deep learning based MI EEG signal classification methods.  
\subsection{Machine Learning}

Machine learning algorithms are widely used in classifying the Motor Imagery (MI) EEG signals. A prominently used feature extraction method known as CSP \cite{gaur2021sliding} with classifier as  SVM, employs matrix diagonalization to identify optimal spatial filters. These filters enhance the variance differentiation of two MI task signal types, resulting in a more discriminative feature vector. The subsequent SVM classification benefits from this enhanced feature set. In comparison, the FBCSP \cite{ang2008filter} technique focuses on extracting features from each frequency band, surpassing the effectiveness of the CSP method. Despite their effectiveness in decoding EEG data, these methods requires manual feature extraction and rely heavily on the researcher's expertise. In contrast, our proposed algorithm eliminates the need for intricate preprocessing and manual feature extraction. By utilizing raw data with complete information, our approach achieving better classification accuracy. This stands in contrast to the limitations faced by machine learning methods, which encounter challenges in achieving further enhancements beyond a certain point.

\subsection{Deep Learning}

Recently, deep learning \cite{lawhern2018eegnet,zhang2020motor,dai2020hs, lotey2022cross,chakladar2022cognitive,keserwani2022robust,kaushik2023motor} has opened several opportunities for EEG-based MI classification problems. % , 
In \cite{lawhern2018eegnet}, a CNN-based EEG classification architecture named EEGNet was introduced that can give promising results on a wide range of BCI paradigms. The EEGNet is a CNN-based network that uses depth-wise and separable convolutions to extract intricate features of multi-domain EEG signals. This work tries to give a generic classification architecture that can efficiently classify EEG signals of different BCI paradigms. The classification efficiency has been validated by applying EEGNet on four domains of EEG signals, namely, P300 event-related potential, feedback error-related negativity, movement-related cortical potential, and sensory-motor rhythm \cite{tangermann2012review}. 
However, this network is shallow and struggles to learn the deep and relevant features.

In CNN, a specific-sized kernel convolves over the entire signal. However, a specific-sized kernel may not extract meaningful features for MI-EEG signal of each individual. Dai \emph{et al.} \cite{dai2020hs} proposed a solution to this problem by using three different scales of kernels in the CNN model, referred to as hybrid scaled CNN (HS-CNN).
% One of the biggest problems with EEG data is the scarcity of correctly recorded high-quality data. The authors have solved this by using a data augmentation technique for MI-EEG signal data. 
Furthermore, Roots \emph{et al.} \cite{roots2020fusion} proposed a multi-branch 2D CNN-based fusion network that used the fusion of outputs obtained from different branches of the network, where each branch has varying hyper-parameters. 

EEG signals are highly subject-dependent, and data from different individuals can be processed efficiently by varying the kernel size. The data from all the network branches can be concatenated and fed to a fully connected layer before classification. The research direction of multi-scale CNN was explored in the form of mobile inception EEGNet (MI-EEGNet), a CNN proposed by Riyad \emph{et al.} \cite{riyad2021mi}. It used a sequence of separable and depth-wise convolutions and an inception block inspired by the computer vision models: Inception and Xception, with different levels of hyper-parameters to extract the most suitable features of EEG signals. Using separable and depth-wise convolutions in place of simple convolutions helps to reduce the number of parameters. Different kernel sizes in the inception block helps to choose various features that act as a countermeasure for overfitting.

\begin{figure*}[!t]
\centering
 \includegraphics[width=\textwidth]{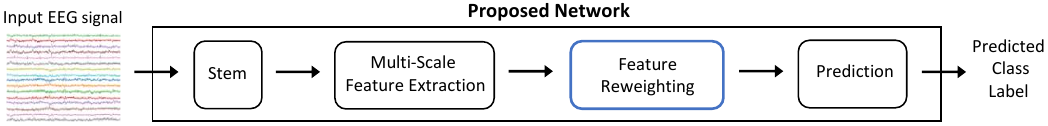}
  \caption{\textcolor{black}{The proposed motor imagery EEG signal classification framework. A 2-D input EEG signal is fed to the proposed network, which predicts the class of the input signal. The novel feature reweighting module is shown in blue box. }}
  \label{fig:proposed_framework}
\end{figure*}

Li \emph{et al.}\cite{li2021temporal} introduced a squeeze and excitation approach-based network that includes CNN-based modules to extract temporal and spectral features and fuses these two features with the help of squeeze-excitation-convolution. Schirrmeister \emph{et al.} \cite{schirrmeister2017deep} introduced two CNN-based architectures named DeepConvNet, that has more convolution layers, and ShallowConvNet, that has fewer convolution layers. Due to the low depth of the architecture, ShallowConvNet proved to perform better for the dataset with fewer samples, as a shallower network helped to avoid overfitting. On the other hand, DeepConvNet performs better for datasets with a large number of data samples, as a huge volume of training data helps to learn a deep learning model efficiently. All approaches mentioned above either use a shallower neural network that cannot efficiently learn the features from MI-EEG data or suffers from irrelevant information across temporal and channel features maps. Miao et al.\cite{miao2023lmda} introduced a lightweight channel and depth attention-based convolutional neural network that is more suitable for real-time BCI applications. However, the model's performance is compromised due to the presence of considerable irrelevant and redundant information in the data.

\textcolor{black}{Our method is different from the approaches mentioned above. It inherently filters the deep features compared to the conventional way of directly passing the computed features from one CNN layer to the next. This work introduces the \textit{feature reweighting} approach that suppresses the less contributing information across temporal and channel feature maps through temporal and channel feature score modules. These scores are then combined using the score fusion module. This module gives a score of feature reweighting that reweight the feature maps of the network in order to reduce the impact of irrelevant information and increase classification performance. Feature reweighting significantly improves classification performance by inherently suppressing the less contributing features across temporal and channel dimensions of the CNN feature maps of MI-EEG signals.}

\section{Method}
\label{sec:proposed_method}
\textcolor{black}{In this section, the framework for EEG signal classification and details of proposed network are discussed. The proposed framework is shown in Figure \ref{fig:proposed_framework}. The input EEG signal is passed as an input to the proposed network, and proposed network classifies the input signal.} Firstly, four modules of the proposed network (stem, multi-scale feature extraction, feature reweighting, and prediction) are discussed. Thereafter, submodules of feature reweighting module are discussed in detail. Furthermore, the procedure of the network training is detailed.

\subsection{Proposed Network} 

The input EEG signals are defined as $E = \{(X_i ,Y_i)|i = 1, 2, . . . , N\}$, where $X_i \in \texttt{R}^{C\times W}$ is a 2D array representing the $i^{th}$ EEG signal, $Y_i \in \{1,2,...M\}$ is the corresponding motor imagery class, $N$ is the total number of EEG signals and $M$ represents the total classes of EEG signals. Here, $C$ and $W$ represent EEG electrodes and temporal data points of EEG signals, respectively. The proposed network consists of four modules: stem, multi-scale feature extraction (MFE), feature reweighting (FR), and prediction. The details of these modules are given as follows:

\begin{figure}[!t]
\centering
 \includegraphics[width=\linewidth]{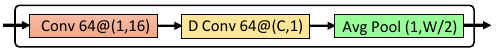}
  \caption{\textcolor{black}{Stem module architecture. Here, 'D conv': Depth-wise convolution layer, K@(k1,k2): K number of convolution kernels and kernel size of (k1,k2), 'Avg Pool (a,b)': average pool layer with output size of (a,b).}}
  \label{fig:stem_module}
\end{figure}

\subsubsection{Stem Module}
The first module of the proposed network is 'Stem' and is similar to the first block of EEGNet \cite{lawhern2018eegnet}. At first, a convolution is applied to the input EEG signal to mix the temporal information, followed by a batch normalization layer. Afterward, a depth-wise convolution is applied to combine the EEG channels into a single channel and filter the spatial dependencies of the input signal. Exponential linear unit (ELU) \cite{clevert2015fast} is an activation function, and it is smooth and differentiable for all values. Therefore, ELU is applied after depth-wise convolution, followed by the dropout and adaptive average pooling \cite{rahimian2020xceptiontime} layer. This module is shown in Figure \ref{fig:stem_module}.

\subsubsection{Multi-scale Feature Extraction Module}
\begin{figure}[!t]
\centering
 \includegraphics[width=\linewidth]{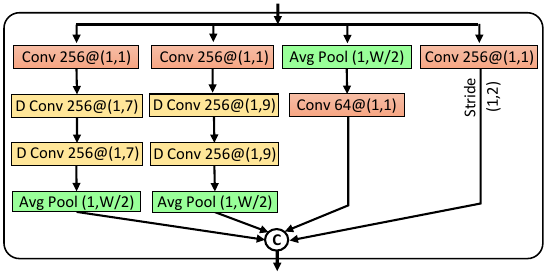}
  \caption{\textcolor{black}{Multi-scale Feature Extraction (MFE) module architecture. Here, 'D conv': Depth-wise convolution layer, K@(k1,k2): K number of convolution kernels and kernel size of (k1,k2), 'Avg Pool (a,b)': average pool layer with output size of (a,b) and $\copyright$: concatenation.}}
  \label{fig:mfe_module}
\end{figure}

From the literature \cite{dai2020hs}, it has been found that the effective convolution kernel size differs for different subjects. The reason is the physiological variations in the EEG signals that are subject-dependent and time-dependent \cite{nijboer2010influence}. Motivated by the finding of \cite{dai2020hs}, the multi-scale kernel approach has been used via the Inception module \cite{szegedy2015going}. This module extracts temporal information on multiple scales through multiple parallel branches that use different convolution kernel sizes. The temporal information on multiple scales helps to increase the EEG signal performance irrespective of the signal variations of each subject. Dropout layer is applied between depth-wise convolution layers to reduce overfitting. This module is depicted in Figure \ref{fig:mfe_module}.

\subsubsection{Feature Reweighting Module}

\begin{figure}[!t]
\centering
 \includegraphics[width=\linewidth]{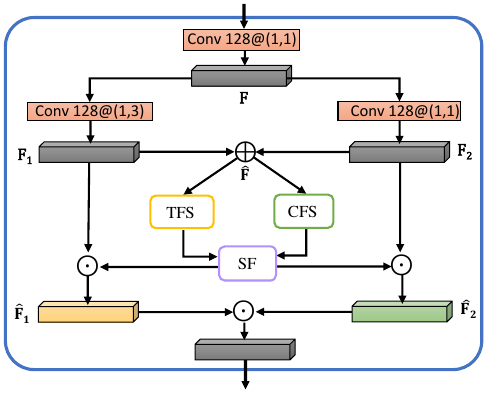}
  \caption{\textcolor{black}{The proposed Feature Reweighting (FR) module architecture. Here, 'Conv': Convolution layer, K@(k1,k2): K number of convolution kernels and kernel size of (k1,k2). Grey, yellow and green cuboids: intermediate feature maps, $\bigoplus$: addition, and $\bigodot$: dot product.}}
  \label{fig:fr_module}
\end{figure}

In the feature reweighting (FR) module depicted in Figure \ref{fig:fr_module}, firstly, the output features from the multi-scale feature extraction (MFE) module are combined via a stack of convolution and ELU activation for mixing the information among various scales of MFE module. The subsequent layers of FR module gives a relevance score by computing the score over temporal and channel dimensions of the feature maps. For computing the temporal and channel scores, two convolution operations are performed on the same feature maps $\mathbf{F}$ to obtain two different feature maps $\mathbf{F_1}$ and $\mathbf{F_2}$. A convolution operation with $1\times3$ kernel size is applied to mix the temporal information of $\mathbf{F}$ feature maps to produce  $\mathbf{F_1}$ feature maps. Similarly, a convolution operation with $1\times1$ kernel size is applied to mix the channels of $\mathbf{F}$ feature maps to produce $\mathbf{F_2}$ feature maps. The output feature maps ($\mathbf{F_1}$ and $\mathbf{F_2}$) are added via element-wise summation to obtain $\hat{\textbf{F}}$ feature maps. 

The temporal and channel feature scores are computed from $\hat{\textbf{F}}$ feature maps through the temporal feature score (TFS) module and channel feature score (CFS) module. TFS and CFS compute the scores ($\mathbf{{S}}_{tfs}$ and $\mathbf{{S}}_{cfs}$) by dropping the irrelevant information in feature maps across temporal and channel dimensions. These scores ($\mathbf{{S}}_{tfs}$ and $\mathbf{{S}}_{cfs}$) are combined using the score fusion (SF) module.
% Individual scores along time and channel dimension of feature maps are combined using score fusion module (SF). 
SF module gives feature reweighting scores $\mathbf{\Hat{S}}_{tfs}$ and $\mathbf{\Hat{S}}_{cfs}$, that are used to filter the feature maps $\mathbf{F_1}$ and $\mathbf{F_2}$, respectively. 
% Feature reweighting is performed to suppress the non-contributing features for efficient classification. 
Inspired by \cite{li2019selective}, the scores ($\mathbf{\Hat{S}}_{tfs}$ and $\mathbf{\Hat{S}}_{cfs}$) are multiplied with the previous features maps ($\mathbf{F_1}$ and $\mathbf{F_2}$) to obtain less irrelevant and more relevant features maps ($\mathbf{{\hat{F}_1}}$ and $\mathbf{{\hat{F}_2}}$). The feature reweighting is performed using the following equations, 
\begin{eqnarray}
\label{eq15.1}
\mathbf{{\hat{F}_1}} = \mathbf{\Hat{S}}_{tfs} \circ \mathbf{F_1}
\end{eqnarray}
\begin{eqnarray}
\label{eq15.2}
\mathbf{{\hat{F}_2}} = \mathbf{\Hat{S}}_{cfs} \circ \mathbf{F_2}
\end{eqnarray}
where $\circ$ is a Hadamard product.

The systematic connection of these sub-modules (TFS, CFS and SF) is shown in Figure \ref{fig:fr_module}. These sub-modules are further discussed in detail in Section \ref{sec1}.

\begin{figure}[!t]
\centering
 \includegraphics[width=\linewidth]{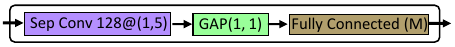}
  \caption{\textcolor{black}{Prediction module architecture. Here, 'Sep conv': Separable convolution layer, K@(k1,k2): K number of convolution kernels and kernel size of (k1,k2), GAP: Global Average Pooling layer, M: number of classes. }}
  \label{fig:prediction_module}
\end{figure}

\subsubsection{Prediction Module}
The features extracted from the feature reweighting module are passed to the prediction module to predict the category of the input MI-EEG signal. Batch normalization and ELU activation applied on the output of FR module, then separable convolution with dropout layer is applied. Thereafter, inspired by \cite{iandola2016squeezenet}, global average pooling layer \cite{rahimian2020xceptiontime} has been included before the classifier. It makes the architecture context independent of the signal length. Finally, a fully connected layer is stacked for classification into M categories. The architecture of the prediction module is depicted in Figure \ref{fig:prediction_module}.

% \begin{landscape}
\begin{figure*}[!t]
\centering
 \includegraphics[width=\linewidth]{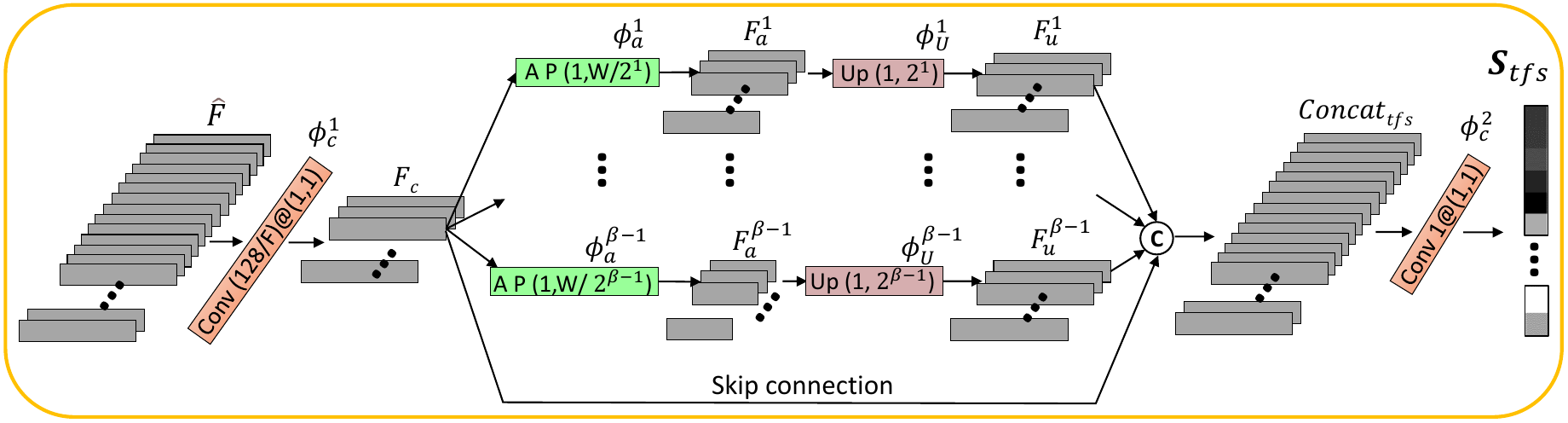}
  \caption{\textcolor{black}{Temporal Feature Score (TFS) Module. Here, Grey blocks: intermediate feature maps, 'Conv K@(k1,k2)': Convolution layer with K kernels and (k1,k2) kernel size, 'A P (a,b)': Average pool layer with output size of (a,b), 'Up (x,y)': Upsample the input features by scale of (x,y), $\copyright$: concatenation of the input feature maps.} }
  \label{fig:tfs_module}
\end{figure*}     
% \end{landscape}

% \vspace{-0.25cm}
\subsection{Feature Reweighting Submodules} 
\label{sec1}

\begin{figure*}[!t]
\centering
 \includegraphics[width=\linewidth]{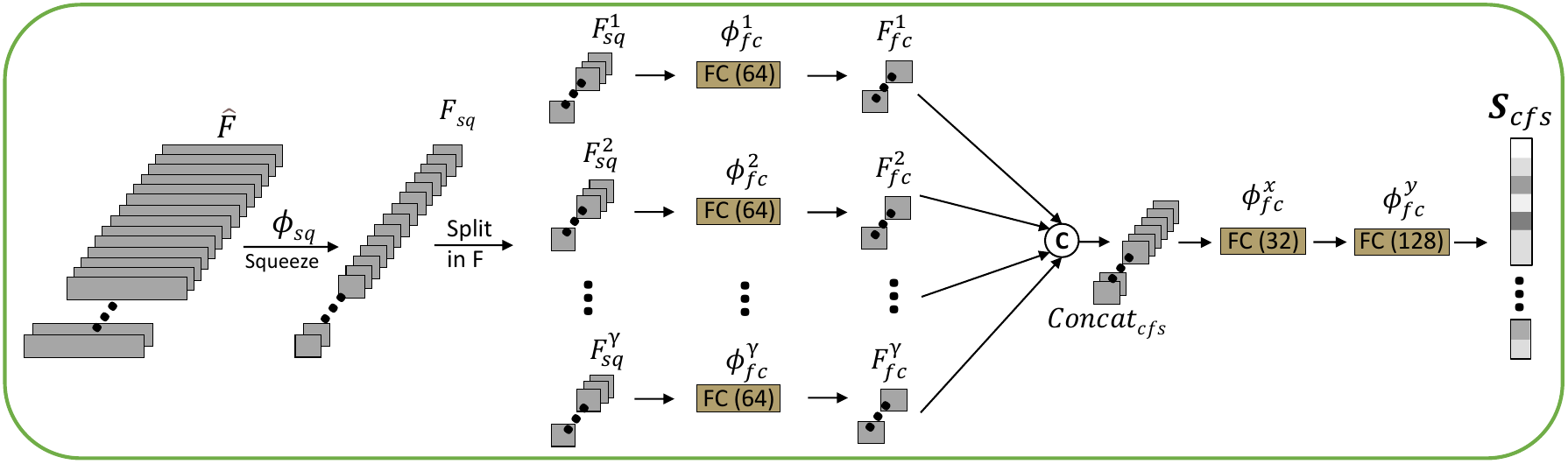}
  \caption{\textcolor{black}{Channel Feature Score (CFS) Module.  Here, Grey blocks: intermediate feature maps, 'FC (M)': Fully connected layer with M neurons, $\copyright$: concatenation of the input feature maps.} }
  \label{fig:cfs_module}
\end{figure*}

\begin{figure}[!t]
\centering
 \includegraphics[width=\linewidth]{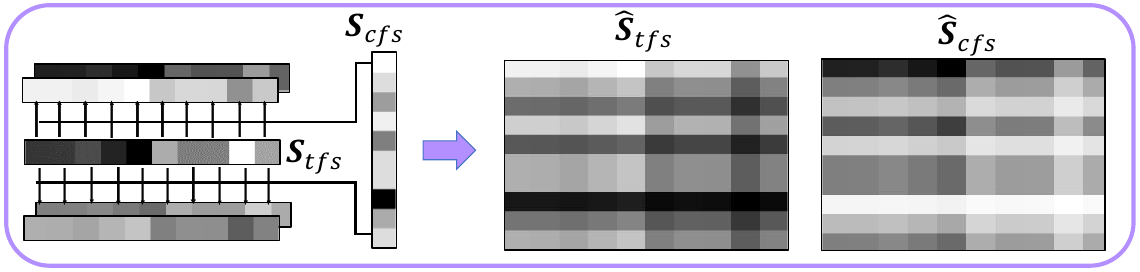}
  \caption{\textcolor{black}{Score Fusion (SF) Module. Temporal feature score ($S_{tfs}$) and channel feature score ($S_{cfs}$) is fused and the output score is used to reweight the features.} }
  \label{fig:sf_module}
\end{figure}

The sub-modules of the feature reweighting module, named as temporal feature score module, channel feature score module, and score fusion module, are shown in Figure \ref{fig:tfs_module}, \ref{fig:cfs_module} and \ref{fig:sf_module}, respectively. The details of these sub-modules are covered in the following subsections: 

% \vspace{0.2cm}
\subsubsection{ \textit{Temporal Feature Score Module:} }
The EEG data is a high-resolution temporal sequence of brain signals, and the EEG signal often gets mixed with other physiological and muscular information \cite{jiang2019removal}. The relevant temporal information mixed with non-important information limits the MI-EEG classification performance. To handle it, the temporal feature score module intelligently learns the score of feature relevance, such that the relevant temporal features get a high score and irrelevant features get a low score.

% filter the irrelevant temporal information and suppress the non-contributing features. 

% \textcolor{black}{This temporal feature score module produces a score through the sequence of operations shown in Figure \ref{fig:idea}. This score value ranges from 0 to 1 and is used to provide weightage to the features. Higher value of this feature score means the features are more important.} 

% \begin{eqnarray}
% \label{eq1}
% \phi_{Conv}(F)=\sum_{c=1}^{C}\sum_{s=1}^{S}f_{c}^{s}*x^{s}
% \end{eqnarray}
In this module, at first, a convolution operation ($\phi_c^1$) is applied to the input feature maps $\hat{\textbf{F}}$ using the following equation,
\begin{eqnarray}
\label{eq2}
\mathbf{F}_{c} = ELU(\phi_{c}^1(\hat{\mathbf{F}}))
\end{eqnarray} 
where ELU is the activation function. The output feature maps ($\textbf{F}_{c}$) are then adaptive average pooled to  different scales ($s_i$) . The number of scales are chosen as temporal scale factor ($\beta$). The scales are given by the following equation,
\begin{eqnarray}
\label{eq3.0}
s_i = 
\begin{matrix}
\alpha .2^{i-1} , & i=0,1,2, ... ,\beta-1
\end{matrix}
\end{eqnarray}
where $\alpha$ is taken as $2$,
% $\beta$ is taken as $4$,
and $s_i$ determines the scale of adaptive average pool. A skip connection is given by $s_0 = 1$ scale, where the input features ($\textbf{F}_c$) are directly passed ahead. 

Adaptive average pooling operation extracts the features at different temporal lengths. The adaptive average pooling ($\phi_a^i$) is performed using the following equation,
% \begin{eqnarray}
% \label{eq3}
% \mathbf{F}_{A}=\phi_A^j(\mathbf{F}_{CE}) = \left\{\begin{matrix}
%  & A(\mathbf{F}_{CE}, i)      &i=1\\ 
%  & A(\mathbf{F}_{CE}, \alpha .2^{i-1})     &i>1
% \end{matrix}\right.
% \end{eqnarray}
\begin{eqnarray}
\label{eq3}
\mathbf{F}_a^i=
\begin{matrix}
\phi_a^i(\mathbf{F}_{c} , s_i), & i=1,2, ... ,\beta-1
\end{matrix}
\end{eqnarray}

The features maps ($\textbf{F}_{a}^{i}$) are then upsampled to the original temporal length to re-scale the relevant features. The scale of upsampling is $s_i$, i.e., same as equation (\ref{eq3.0}). Upsampling operation ($\phi_u^i$) depicted in following equation,
% \begin{eqnarray}
% \label{eq4}
% \mathbf{F}_{U}=\phi_U^j(\mathbf{F}_{A}) = \left\{\begin{matrix}
%  & U(\mathbf{F}_{A}, i)      &i=1\\ 
%  & U(\mathbf{F}_{A}, \alpha .2^{i-1})   &i>1
% \end{matrix}\right.
% \end{eqnarray}
\begin{eqnarray}
\label{eq4}
\mathbf{F}_u^i= 
\begin{matrix}
\phi_u^i(\mathbf{F}_a^i, s_i)    &i=1,2, ... ,\beta-1
\end{matrix}
\end{eqnarray}

% \textcolor{black}{In the above equation, $\alpha$ is 2.5 and $s_i$ determines the scale of upsampling operation.}
All feature maps ($\textbf{F}_u^i$) obtained from the $\phi_u^i$ upsampling operation are concatenated with $\textbf{F}_c$ using the following equation,
% All of the output feature maps are then concatenated with the help of equation (\ref{eq5}):
\begin{eqnarray}
\label{eq5}
\mathbf{Concat}_{tfs}=[\mathbf{F}_u^1 | ... | \mathbf{F}_u^{\beta-1} |  \mathbf{F}_c]
\end{eqnarray}
% where $\textbf{F}_U^i$ is the $i^{th}$ feature maps obtained from the $\phi_U^i$ upsampling operation. 

\noindent where [$x|y$] represents the concatenation of two feature maps $x$ and $y$. A skip connection of input features $\mathbf{F_c}$ shown in Figure \ref{fig:tfs_module}.
% Three scales ($s_1$, ...,  $s_{\beta-1}$) from equation (\ref{eq3}) and 
% (a) gives a total of four (4) scales. This scale factor is denoted by the time scale factor ($\beta$). 
% In this work, value of $\beta$ is taken as 4. 
A convolution operation ($\phi_{c}^2$) is then performed on the concatenated feature maps to obtain a one-dimensional temporal feature score ($\mathbf{S}_{tfs} \in \texttt{R}^{T}$, $T$ is temporal length). The following equation derives the temporal feature score ($\textbf{S}_{tfs}$),
% \vspace{-0.5cm}
\begin{eqnarray}
\label{eq6}
\mathbf{S}_{tfs} = \phi_{c}^2(\mathbf{Concat}_{tfs})
\end{eqnarray}
% This last sequence of operations gives the temporal feature gating score ($\mathbf{S}_{tgm}$) as an output that is of dimension $T$. 
% \textcolor{black}{This score acts as one of the input to the reweighting module. It is used to reweight the input feature maps temporally.} 

\subsubsection{\textit{Channel Feature Score Module}} 
The first step to compute the channel feature score is to squeeze the input feature maps $\mathbf{\hat{F}}$ obtained from the previous operation. This step is performed to explicitly select the relevant features that correspond to the respective feature map channel. 
The squeeze operation ($\phi_{sq}$) is defined as follows,
\begin{eqnarray}
\label{eq7.1}
\mathbf{F}_{sq} = \phi_{sq}(\mathbf{\hat{F}}) = \frac{1}{h\times w}\sum_{i=1}^{w}\sum_{j=1}^{h}\mathbf{\hat{F}}(i,j)
\end{eqnarray}

\noindent where $\mathbf{F}_{sq}$ is the squeezed feature maps. $h$ and $w$ are the height and width of $\mathbf{\hat{F}}$ feature maps, respectively.

For efficient channel feature extraction, inspiration has been taken from \cite{zhang2018shufflenet}. The feature maps can be split into sets of multiple feature maps and then transformed separately for better and relevant feature extraction \cite{zhang2018shufflenet}. This splitting is done through the channel split factor ($\gamma$).  
% Inspired from the \cite{zhang2018shufflenet}, for efficient channel-wise feature extraction, the feature maps are split into multiple feature maps. This splitting of feature maps is done by channel split factor (CSF).
% In this work, the value of $\gamma$ is taken as $4$. 
The split operation gives $\gamma$ sets of feature maps ($\mathbf{F}_{sq}^1$, ..., and $\mathbf{F}_{sq}^\gamma$). %This operation is similar to the \cite{zhang2018shufflenet}, but there is a difference that shuffling of the feature maps across the feature sets is not performed in this work. 

Thereafter, for each $\mathbf{F}_{sq}^i \in \texttt{R}^{s}$, a fully connected layer operation ($\phi_{fc}^i$) is performed to reduce the feature maps size to half i.e. $\mathbf{F}_{fc}^i \in \texttt{R}^{\frac{s}{2}}$ using the following equation,

% these sets of feature maps are fed to different fully connected layer operations to reduce the number of the feature maps by the scale of $2$. The operation is computed using equation (\ref{eq7}),
% \begin{eqnarray}
% \label{eq7}
% \mathbf{F}_{L}^{i}=\mathbf{W*x} + \mathbf{b}
% \end{eqnarray}
\begin{eqnarray}
\label{eq7}
\mathbf{F}_{fc}^{i} = 
\begin{matrix}
ReLU(\phi_{fc}^{i}(\mathbf{F}_{sq}^i))    &i=1,2,..., \gamma
\end{matrix}
\end{eqnarray}

\noindent where Rectified Linear Unit (ReLU) is used as the activation function of the fully connected layer. ReLU is chosen for its ability to introduce sparsity in the gradient, thereby aiding in the reduction of overfitting and to suppress the irrelevant information present in the feature maps. All of the output feature maps ($\mathbf{F}_{fc}^1$, $\mathbf{F}_{fc}^2$, ...,  $\mathbf{F}_{fc}^\gamma$) of the fully connected layer operation $\phi_{fc}^i$ are then concatenated with the help of following equation,
\begin{eqnarray}
\label{eq8}
\mathbf{Concat}_{cfs}=[\mathbf{F}_{fc}^{1}|\mathbf{F}_{fc}^{2}|...|\mathbf{F}_{fc}^{\gamma}]
\end{eqnarray}

\noindent where [$x|y$] represents the concatenation of two feature maps $x$ and $y$. In this work, the value of $\gamma$ is taken as $4$. 
% where $\mathbf{F}_{fc}^i$ are the $i^{th}$ feature maps obtained from the $\phi_{fc}^i$ fully connected layer operation.
% \begin{eqnarray}
% \label{eq9}
% \phi_{P}=\phi_{L}^{'}(\delta(\phi_{L}(x)))
% \end{eqnarray}
The channel feature score ($\mathbf{S}_{cfs} \in \texttt{R}^P$, P is no. of channels) is then obtained using the following equation,
\begin{eqnarray}
\label{eq10.0}
\mathbf{S}_{cfs}=\phi_{{fc}}^y(ReLU(\phi_{fc}^x(\mathbf{Concat}_{cfs})))
\end{eqnarray}
% \begin{eqnarray}
% \label{eq10.1}
% \phi_{P}(\mathbf{X})=\phi_{{fc}}^6(\delta(\phi_{fc}^5(\mathbf{X})))
% \end{eqnarray}
where $\phi_{{fc}^x}$ and $\phi_{{fc}^y}$ are fully connected layer operations that are used for the projection of the feature maps to the P dimension.

\subsubsection{\textit{Score Fusion Module} }
The temporal feature score ($\mathbf{S}_{tfs}$) and channel feature score ($\mathbf{S}_{cfs}$)  
% are of dimension $N$ and $T$, respectively. 
% The feature reweighting map is obtained by equation (\ref{eq14}),
% \begin{eqnarray}
% \label{eq14}
% \mathbf{\beta} = \mathbf{R}_{cgm} \times \mathbf{R}_{tgm}
% \end{eqnarray}
% where $\times$ is the matrix multiplication operation. The dimension of the $\mathbf{\beta}$ is $(N \times T)$. 
% These score 
are fused to compute the feature reweighting scores ($\mathbf{\Hat{S}}_{tfs}$ and $\mathbf{\Hat{S}}_{cfs}$). The fusion of the scores is carried out through the following equations,

\begin{eqnarray}
\label{eq11.0}
\mathbf{\Hat{S}}_{tfs}^{(i,j)}=\frac{e^{\mathbf{S}_{tfs}^{(i,j)}}}{e^{\mathbf{S}_{tfs}^{(i,j)}}+e^{\mathbf{S}_{cfs}^{(i,j)}}}
\end{eqnarray}

\begin{eqnarray}
\label{eq11.1}
\mathbf{\Hat{S}}_{cfs}^{(i,j)}=\frac{e^{\mathbf{S}_{cfs}^{(i,j)}}}{e^{\mathbf{S}_{tfs}^{(i,j)}}+e^{\mathbf{S}_{cfs}^{(i,j)}}}
\end{eqnarray}

$\mathbf{\Hat{S}}_{tfs}$ and $\mathbf{\Hat{S}}_{cfs}$ are used thereafter in equations Eq. \ref{eq15.1} and Eq. \ref{eq15.2} to reweight features $\mathbf{F}_1$ and $\mathbf{F}_2$.

%%%%%%%%%%%%%%%%%%%%%%%%%%%%%%%%%%%%%%%%%%%%%%
\newcommand{\nosemic}{\renewcommand{\@endalgocfline}{\relax}}% Drop semi-colon ;
\newcommand{\dosemic}{\renewcommand{\@endalgocfline}{\algocf@endline}}% Reinstate semi-colon ;
\newcommand{\pushline}{\Indp}% Indent
\newcommand{\popline}{\Indm\dosemic}% Undent
\let\oldnl\nl% Store \nl in \oldnl
\newcommand{\nonl}{\renewcommand{\nl}{\let\nl\oldnl}}% Remove line number for one line
\begin{algorithm}[!ht]
\begin{normalsize}

  \SetAlgoLined
  \SetKwInput{KwInput}{Input}                % Set the Input
  \SetKwInput{KwOutput}{Output}              % set the Output
  \SetKwInput{KwProcedure}{Procedure}              % set the Output
   \KwInput{Train set ($E_{train}$), validation set ($E_{val}$), training epochs ($N$), learning rate ($\eta$), factor ($f$), patience ($\tau$), model $M(\cdot)$}
   % , weight decay ($\delta$)
  \KwOutput{Trained proposed model $M(\theta^{N})$}
  % \KwProcedure{}  % \nonl Initialize one batch data $X_i$ where $i=1,2,...,b$\;
  \nonl Initialize model parameters $\theta^{0}$ $\gets$ Xavier method,  
 % $\eta \gets 10^{-3}$, $\delta \gets 
 % 10^{-4}$,$f \gets  0.1$  \\
 % $N \gets  200$, $\tau \gets  25$  \\
  epoch counter $k \gets  1$,  patience counter $a \gets 0$, minimum validation loss $V^{'}_{loss} \gets 0$, local validation loss $V_{loss} \gets 0$\\
  % $k \gets  1$, $V^{'}_{loss} \gets 0$,  $a \gets 0$\\
\While{($k \le N$)}{  %\textbf{and} $a<\tau$
    % k++\;
    % $T_{loss} \gets 0$\\
    $V_{loss} \gets 0$\\ 
    \For{each mini-batch $B^i$ in $E_{train}$}{
    $(D_i , L_i) \gets B^i$\\
    $p_i = M(\theta^{i}_k, D_i)$\\ %Generate class probability 
    $l_{train} \gets$ CEL($p_i$,$L_i$)\\
    % + $\delta \times R(\theta^{k})$\\
    % $T_{loss} \gets T_{loss} + l_{train}$ \\ 
    %Backpropogate the loss in $Net(\theta^{k})$\\
    $\theta^{i+1}_{k} \gets  \theta^{i}_k - \eta \times \nabla l_{train}$ \\   %Update weights to 
    }
    \For{each mini-batch $B^i$ in $E_{val}$}{
    $(D_i , L_i) \gets B^i$\\
    $p_i = M(\theta^{i}_k, D_i)$\\ %Generate class probability 
    $l_{val} \gets$ CEL($p_i$,$L_i$)\\
    % + $\delta \times R(\theta^{k})$\\
    $V_{loss} \gets V_{loss} + l_{val}$ \\ 
    }
    \If{$V_{loss} < V^{'}_{loss}$ }{
    $V^{'}_{loss} \gets V_{loss}$
    % \If{$V_{loss} < l_{val}$ }{
    % $V_{loss} \gets l_{val}$ \\
    \Else{
    $a \gets a + 1$\\
    }}
    \If{$a > \tau$ }{
    $\eta \gets \eta  \times f$, 
    $a \gets 0$
    }
    $k \gets k + 1$\\
    %Calculate loss $J^{k}$ on $E_{train}$ by the equation $->$
    % $l_{train} = -\sum_{i=1}^{M}\hat{y}_{i}log(y_i)$ \;
    % Calculate loss $l_{val}$ on $E_{val}$ by the equation (\ref{eq:12})
    % \For{each item in data}{
    %   Process item\;
    % }
    % Update parameters\;
  }
  \caption{The network training procedure}
  \label{Algo1}
\end{normalsize}
\end{algorithm}
%%%%%%%%%%%%%%%%%%%%%%%%%%%%%%%%%%%%%%%%%%%%

% \vspace{-0.35cm}
\subsection{Network Training Procedure}
The step-by-step optimization process of the proposed network training is illustrated in Algorithm \ref{Algo1}. The whole dataset is split into two sets, i.e., training and testing data. The training data is further split into two sets, i.e., the train set, and the validation set. The proposed model learns the classification of data by minimizing the prominently used cross-entropy loss \cite{lotey2022cross}. The cross-entropy loss function denoted as $CEL(p_i,L_i)$, is given in the following equation, 
\begin{eqnarray}
\label{eq16}
CEL(p_i,L_i) = -\sum_{i=1}^{M}L_{i}log(p_i)
\end{eqnarray}
where M refers to the number of classes.

The loss is computed between model predictions ($p_i$) and actual label ($L_i$) of data ($D_i$). The model parameters ($\theta$) are updated after each minibatch ($B$) according to the gradient of computed training loss ($\nabla$$l_{train}$) and learning rate ($\eta$). Simultaneously, validation loss ($l_{val}$) is computed on the validation set. A widely used updation criteria for learning rate is used in this work. The learning rate of the model is decreased by the factor ($f$) if validation loss doesn't decrease for consecutive $\tau$ epochs. The pseudo-code for gradient computation and learning rate updation criteria is shown in Algorithm \ref{Algo1}.

\section{Experiments and Results}
\label{sec:experiments}

This section discusses the datasets, experimental details, and the various metrics used in this study for validation of the results. It also gives a detailed analysis of the classification results obtained by the proposed method on both of the analyzed motor imagery EEG datasets. Additionally, the classification performance comparison with other state-of-the-art methods is also analyzed in this section. \textcolor{black}{Furthermore, the generalizability of the proposed method for tasks other than motor imagery is validated by evaluating the classification performance on speech imagery and motor movement datasets.}

\subsection{\textcolor{black}{Dataset Description}}
\textcolor{black}{In this work, two publicly available motor imagery EEG datasets have been used to validate the performance and generalizability of the proposed method. The statistical details of the datasets are shown in Table \ref{table:dataset_stats} and their detailed description is as follows:}

\subsubsection{\textcolor{black}{Dataset 1}}
\textcolor{black}{The first dataset used in this work is a motor imagery EEG dataset named 'BCI Competition IV 2a' \cite{tangermann2012review}. It is a collection of motor imagery EEG data of nine healthy participants. The dataset contains EEG signals corresponding to four MI tasks, namely, left-hand, right-hand, both feet, and tongue. All of these tasks are used in this work. The EEG signals were recorded using $22$ EEG electrodes with a sampling rate of $250$ Hz. All $22$ EEG electrodes and the time segment of $0-4$ seconds after the onset of the visual cue are used in this work. To improve the signal-to-noise ratio of the EEG signals, widely accepted artifact removal steps have been used in this study. The data preprocessing steps, including removal of muscular and eye movement artifacts, min-max normalization, and independent component analysis, are consistent with those used in the studies by \cite{schirrmeister2017deep, riyad2021mi}. In the rest of the paper, \textbf{'MI-BCI'} will be used to refer this motor imagery dataset.}

\begin{table}[!t]
\centering
\setlength{\tabcolsep}{3.5pt}
\caption{\textcolor{black}{Statistics of datasets used in this work.}}
\footnotesize

\begin{tabular}{l|ccccc}
\hline
\textbf{Task} & \textbf{Dataset} & \textbf{\#Subjects} & \textbf{\#Channels} & \textbf{\#Trials} & \textbf{\#Classes} \\ \hline
MI & MI-BCI & 9           & 22          & 576       & 4          \\
MI & MI-PYD & 109         & 64          & 90        & 4          \\ 
SI & SI-ASU & 6           & 60          & 300       & 3          \\   
ME & MM-HGD & 14          & 128         & 480       & 4          \\ \hline
\end{tabular}
\label{table:dataset_stats}
    
\end{table}

\subsubsection{\textcolor{black}{Dataset 2}}
\textcolor{black}{The second dataset utilized for this work is a motor imagery dataset from Physionet \cite{goldberger2000physiobank}. This dataset contains four motor imagery tasks, namely, right fist, left fist, both fists and both feet have been taken for analysis. The data is recorded on $64$ EEG electrodes with a sampling rate of $160$ Hz. It contains data from $109$ subjects. All $64$ EEG electrodes and the time segment of $0-4$ seconds after the onset of the visual cue are used in this work. Six subjects ($38$, $88$, $89$, $92$, $100$, and $104$) out of $109$, are dropped from the dataset due to different sampling rates and wrong labeling. Similar to the MI-BCI dataset, the pre-processing steps employed in this study for physionet motor imagery dataset are same as used in the study of \cite{roots2020fusion}.  Henceforth in this paper, \textbf{'MI-PYD'} will be used to refer this motor imagery dataset.}

\subsubsection{\textcolor{black}{Dataset 3}}
\textcolor{black}{For generalizability analysis of the proposed method, a speech imagery EEG dataset is included in this work. This dataset is a publicly available EEG-based speech imagery dataset recorded at Arizona state university (ASU), where participants imagined the speech for words 'in', 'out' and 'up' \cite{nguyen2017inferring}. A total of $6$ subjects performed the short word speech imagery task with sampling rate of $256$ Hz. The number of trials for each class is $100$, that gives total number of $300$ trials. The data was collected through $64$ EEG channels. During preprocessing, a butterworth band pass filter between $8-70$ Hz and a notch filter of $60$ Hz is applied \cite{nguyen2017inferring}. Similar to \cite{nguyen2017inferring}, $60$ EEG channels are chosen for this work. Henceforth in this paper, \textbf{'SI-ASU'} will be used to refer this speech imagery dataset.}

\subsubsection{\textcolor{black}{Dataset 4}}
\textcolor{black}{ A motor movement task based EEG dataset is also included in this work to further analyze the generalizability of the proposed method. 
The standard public high gamma dataset (HGD) has been chosen for this study \cite{schirrmeister2017deep}. HGD comprises EEG data from $128$ EEG electrodes from $14$ healthy subjects. The sampling frequency of this dataset is $250 Hz$. The dataset comes with four category tasks, namely, left-hand movement, right-hand movement, both feet movement and rest. This study uses $44$ electrodes covering the motor cortex, similar to \cite{schirrmeister2017deep}. Henceforth in this paper, \textbf{'MM-HGD'} will be used to refer this motor movement dataset.}

\subsection{Experimental Setup} 
The proposed network has been trained using a Xeon processor with $96$GB RAM and Nvidia Quadro P5000 with $16$GB of GPU RAM. The proposed model has been initialized with Xavier method \cite{glorot2010understanding} and trained for $200$ epochs with Adam optimizer \cite{kingma2014adam} with an initial learning rate of $0.001$ and weight decay of $10^{-4}$. The model has been trained with a batch size of $64$. We utilize the 5-fold cross-validation technique, which requires dividing the dataset into five disjoint folds. During each iteration, one fold serves as the test set while the remaining four folds function as the training set. This process is repeated five times, ensuring that each fold is utilized as the test set once. The performance metrics derived from these iterations are averaged to yield comprehensive evaluation results. Also, the same experimental settings are used for both datasets.

% The learning rate of the optimizer is reduced by a factor of $0.1$ when the validation loss is not decreasing for consecutive $25$ epochs. 
%For the inference, the last epoch weights are considered. 

%  and betas of $(0.5, 0.999)$

% \subsection{Comparison models} 
% will add this section if there is vacant space

% \vspace{-0.3cm}

\subsection{Performance Metrics}
In this work, the metrics used for the evaluation of MI-EEG classification are accuracy \cite{li2021temporal}, f-measure \cite{li2019channel}, and Kappa measure \cite{carletta1996assessing}. 
% The mathematical equations for the these measure are given as following:

% \begin{align}
% Accuracy = \frac{TP}{(TP+TN+FP+FN)}
% \end{align}
% where, TP, FP, TN, FN are the true positive, false positive, true negative and false negative, respectively. 
% \begin{align}
% Precision = \frac{TP}{(TP+FP)}.\label{eq:precision}
% \end{align}
% \begin{align}
% Recall = \frac{TP}{(TP+TN)}.\label{eq:recall}
% \end{align}
% Equation \ref{eq:fmeasure} is obtained using equation \ref{eq:precision} and equation \ref{eq:recall}. It is given as follows:

% \begin{align}
% F-measure = 2.\frac{Precision.Recall}{(Precision+Recall)}.\label{eq:fmeasure}
% \end{align}

The kappa measure can be computed using the following mathematical expression:
\begin{align}
\kappa &= \frac{P_o - P_e}{1 - P_e}.\label{eq:kappa}
\end{align}
where $\kappa$ denoted kappa measure, $P_o$  is the observed agreement and $P_e$ is the expected agreement by chance.

In addition, AUC-ROC curve plots \cite{fawcett2006introduction} are used to measure the degree of separability of the model among various classes. 
Furthermore, the paired t-test method is employed to measure the statistical significance of the proposed method over the compared methods.

\subsection{Classification Performance}

\begin{table}[H]
\centering
\caption{Detailed results of the proposed method on the MI-BCI and MI-PYD datasets.}
\small

\begin{tabular}{l|llcccc} 
\toprule
\textbf{Dataset}&\textbf{Sub} & \textbf{Accuracy} &
\textbf{F-measure} 
% & \textbf{AUC-ROC} 
% & \textbf{Kappa} 
\\
\midrule
\multirow{10}{*}{\textbf{MI-BCI}} &\textbf{A01}    & 85.54 $\pm$ 3.02 & 0.85 $\pm$ 0.03 
% & 0.97 $\pm$ 0.01 
% & 0.81 $\pm$ 0.04 
\\

&\textbf{A02}    & 60.89 $\pm$ 5.38 & 0.61 $\pm$ 0.06 
% & 0.83 $\pm$ 0.02 
% & 0.48 $\pm$ 0.07 
\\

&\textbf{A03}    & 93.75 $\pm$ 1.95 & 0.94 $\pm$ 0.02 
% & 0.99 $\pm$ 0.00 
% & 0.92 $\pm$ 0.03 
\\
 
&\textbf{A04}    & 78.16 $\pm$ 4.12 & 0.78 $\pm$ 0.04 
% & 0.95 $\pm$ 0.01 
% & 0.71 $\pm$ 0.05 
\\

&\textbf{A05}    & 72.67 $\pm$ 3.81 & 0.72 $\pm$ 0.04 
% & 0.92 $\pm$ 0.01 
% & 0.63 $\pm$ 0.05 
\\

&\textbf{A06}    & 69.39 $\pm$ 5.47 & 0.69 $\pm$ 0.05 
% & 0.90 $\pm$ 0.03 
% & 0.59 $\pm$ 0.07 
\\

&\textbf{A07}    & 93.80 $\pm$ 3.25 & 0.94 $\pm$ 0.03 
% & 0.99 $\pm$ 0.00 
% & 0.92 $\pm$ 0.04 
\\

&\textbf{A08}    & 88.22 $\pm$ 3.94 & 0.88 $\pm$ 0.04 
% & 0.99 $\pm$ 0.01 
% & 0.84 $\pm$ 0.05 
\\

&\textbf{A09}    & 83.64 $\pm$ 3.42 & 0.83 $\pm$ 0.04 
% & 0.96 $\pm$ 0.02 
% & 0.78 $\pm$ 0.05 
\\
&\textbf{Avg.}    & 80.67 $\pm$ 11.30 & 0.80 $\pm$ 0.11 
% & 0.96 $\pm$ 0.02 
% & 0.74 $\pm$ 0.15 
\\
 \midrule
 \textbf{MI-PYD} & \textbf{Avg.}  & 95.05 $\pm$ 2.11  & 0.95 $\pm$ 0.02  
 % & 0.93 $\pm$ 0.03
 \\
\bottomrule
\end{tabular}
\label{table:bci_results}    

\end{table}

\begin{figure}[!t]
\centering
 \includegraphics[width=\linewidth]{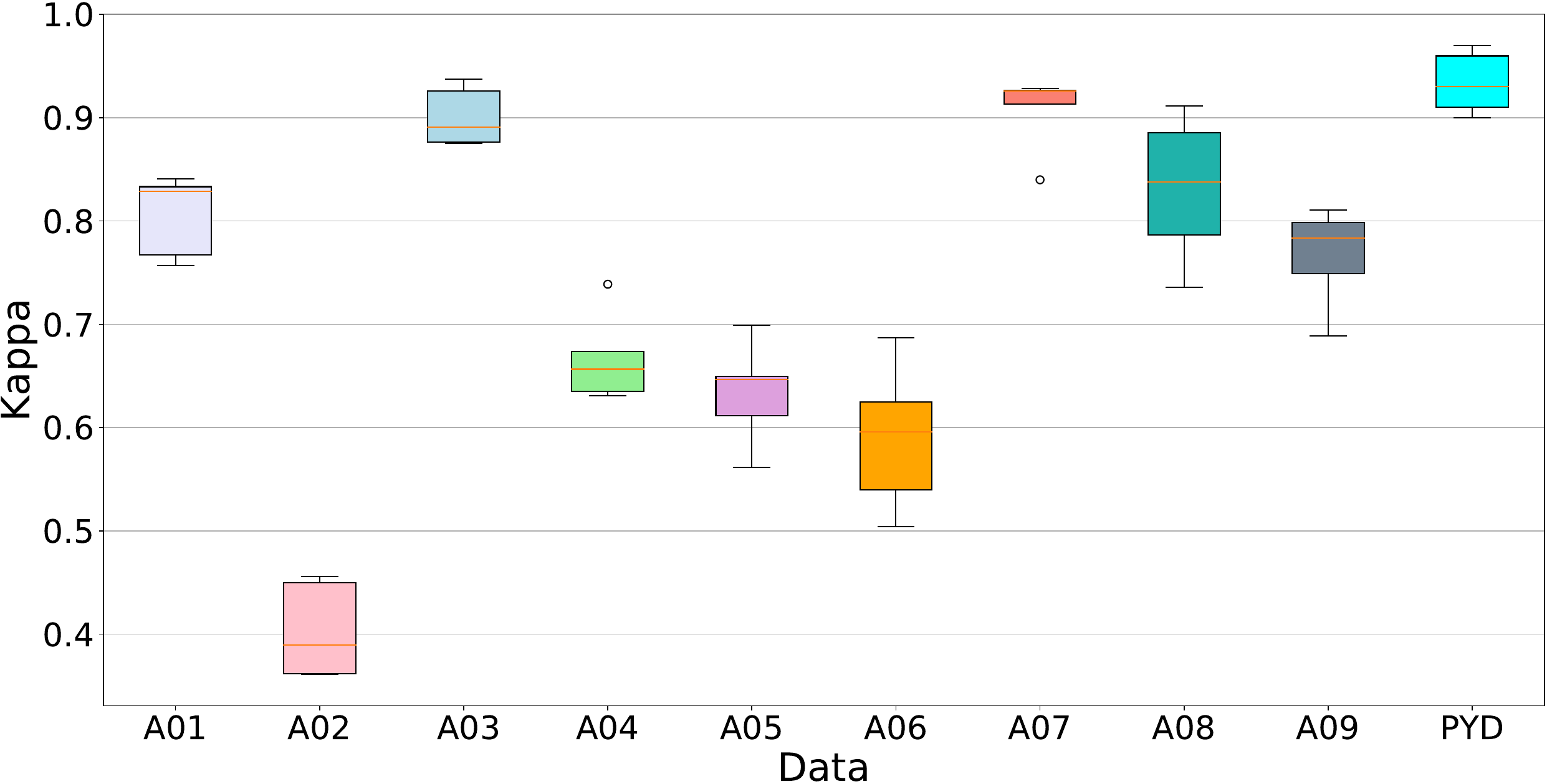}
  \caption{\textcolor{black}{Kappa values of the proposed method for MI-BCI (subjects A01,..., A09) and MI-PYD dataset).} }
  \label{fig:kappa_proposed}
\end{figure}

In this section, the classification performance of the proposed method for MI-BCI and MI-PYD datasets has been discussed in detail. For validation of the generalizability of the proposed method, two types of experiments have been performed. MI-BCI dataset is used to check the subject-wise classification performance as the data samples are less in subject-wise setting. Secondly, MI-PYD dataset dataset contains large number of data samples. Therefore, MI-PYD dataset is used to validate the proposed method's performance over whole dataset. The detailed results of the proposed method for four class MI-EEG tasks of MI-BCI and MI-PYD datasets are summarized in Table \ref{table:bci_results}. The average accuracy of proposed method for the MI-BCI dataset is $80.67\%$. Table \ref{table:bci_results} shows that five subjects (A01, A03, A07, A08 and A09) have an accuracy of more than $80\%$. A similar pattern can be observed for F-measure. It shows the good prediction performance of the proposed method. 

\textcolor{black}{Kappa measure is a statistical measure of agreement between values predicted by the classifier and the actual values. Therefore, in this work, kappa measure is computed to analyze the class separability and classification performance of the proposed method. Figure \ref{fig:kappa_proposed} shows the mean and variance of performance of the proposed method on all nine subjects of MI-BCI dataset (A01,..., A09) and MI-PYD dataset (PYD). It can be observed that subjects A01, A03, A07, and A09 have a mean kappa value higher than $0.8$, which indicates a higher degree of class separability. With only the exception of subject A02, which performed fairly on class separability, all other subjects performed substantially better. This observation suggests the proposed method performs well on class separability, validated through the kappa measure. Furthermore, the kappa measure for MI-PYD dataset (indicated as PYD) is very high, indicating excellent class separability of the proposed method.}

\begin{figure*}[!t]
    \centering
    % \hspace{-0.2cm}
    \begin{subfigure}[t]{0.2\textwidth}  
            \centering 
            \includegraphics[scale=0.052]{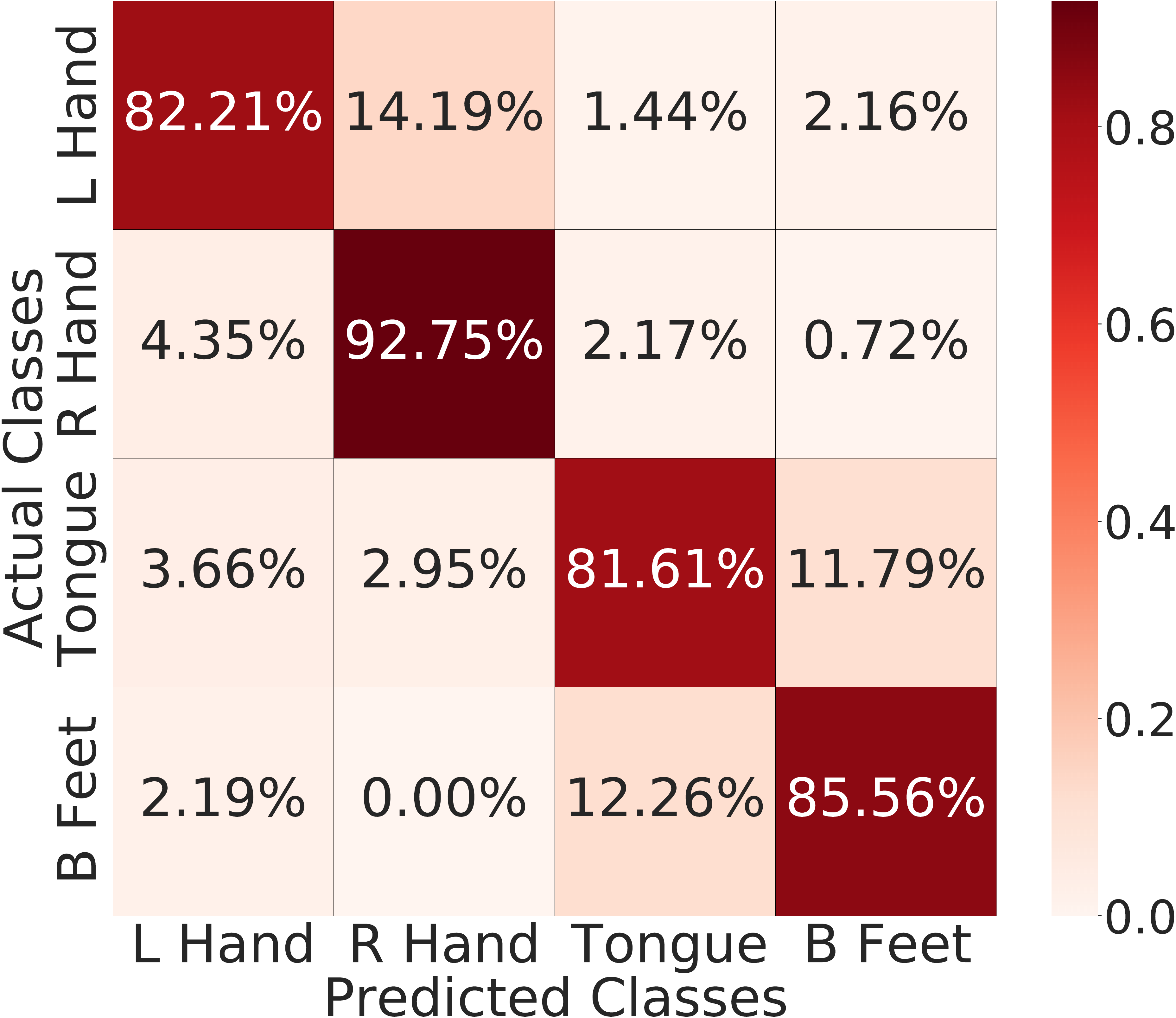}
            \caption[]%
            {{\small A01}}    
            \label{fig:cm_a01}
    \end{subfigure}%
    ~      
    \hspace{-0.3cm}
    \begin{subfigure}[t]{0.2\textwidth}  
            \centering 
            \includegraphics[scale=0.052]{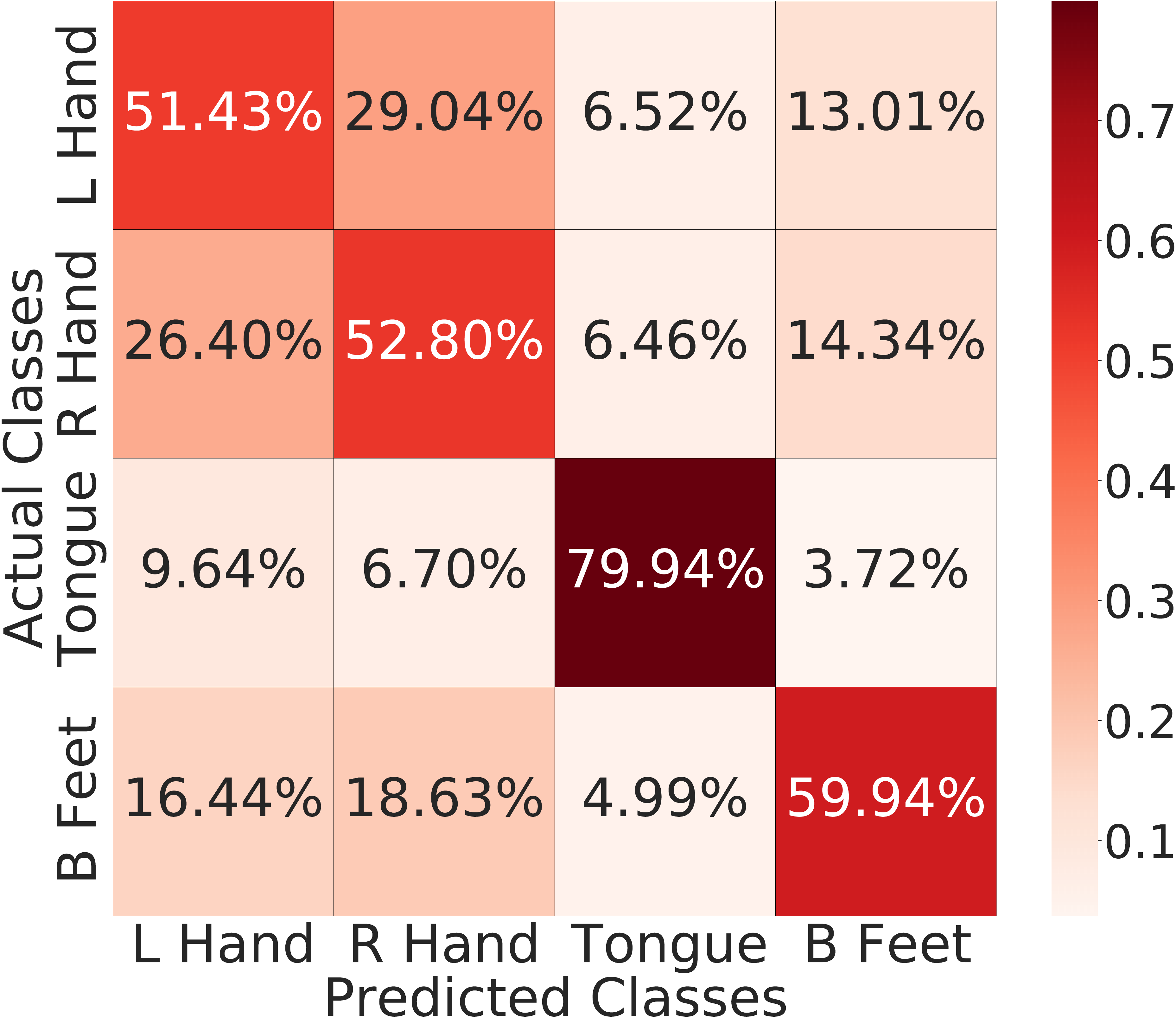}
            \caption[]%
            {{\small A02}}    
            \label{fig:cm_a02}
    \end{subfigure}%
    ~ 
    \hspace{-0.3cm}
    \begin{subfigure}[t]{0.2\textwidth}  
            \centering 
            \includegraphics[scale=0.052]{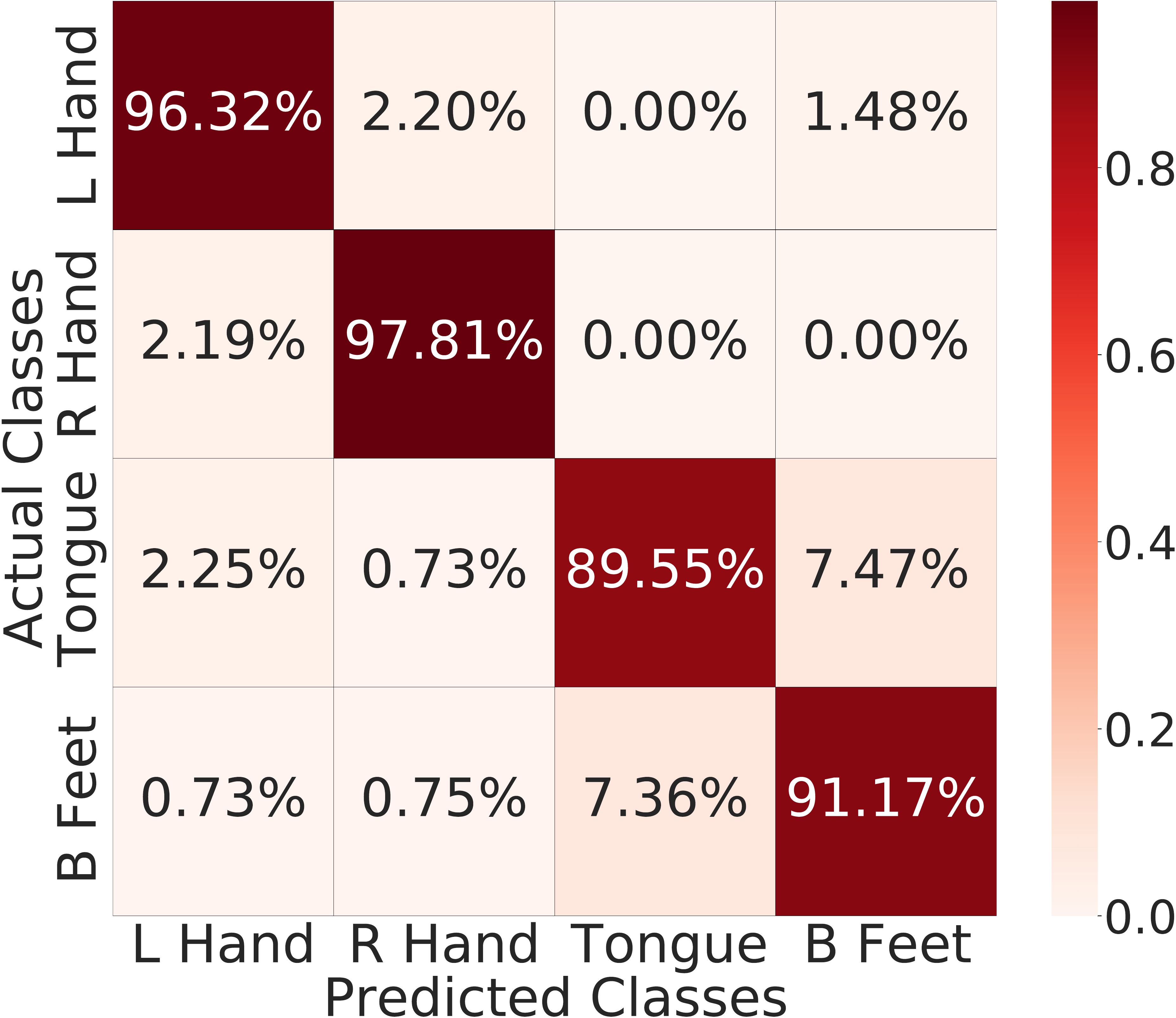}
            \caption[]%
            {{\small A03}}    
            \label{fig:cm_a03}
    \end{subfigure}%
    ~ 
        \hspace{-0.3cm}
        \begin{subfigure}[t]{0.2\textwidth}  
            \centering 
            \includegraphics[scale=0.052]
            {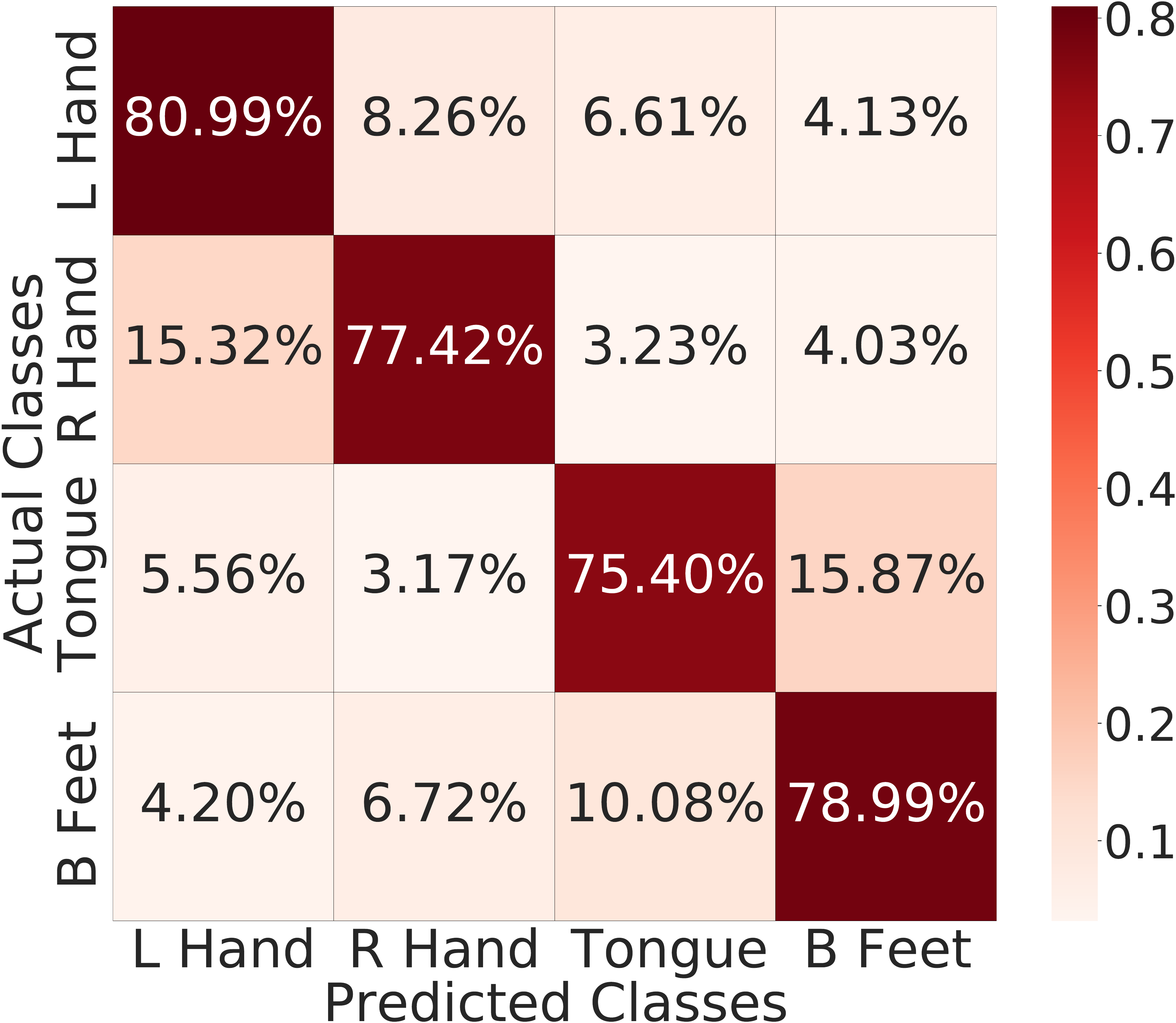}
            % {images/graphs/cm_pyd_average.pdf}
            \caption[]%
            {{\small A04}}%PYD}}    
            \label{fig:cm_a04}
        \end{subfigure}%
    ~
        \hspace{-0.3cm}
       \begin{subfigure}[t]{0.2\textwidth}  
            \centering 
            \includegraphics[scale=0.052]{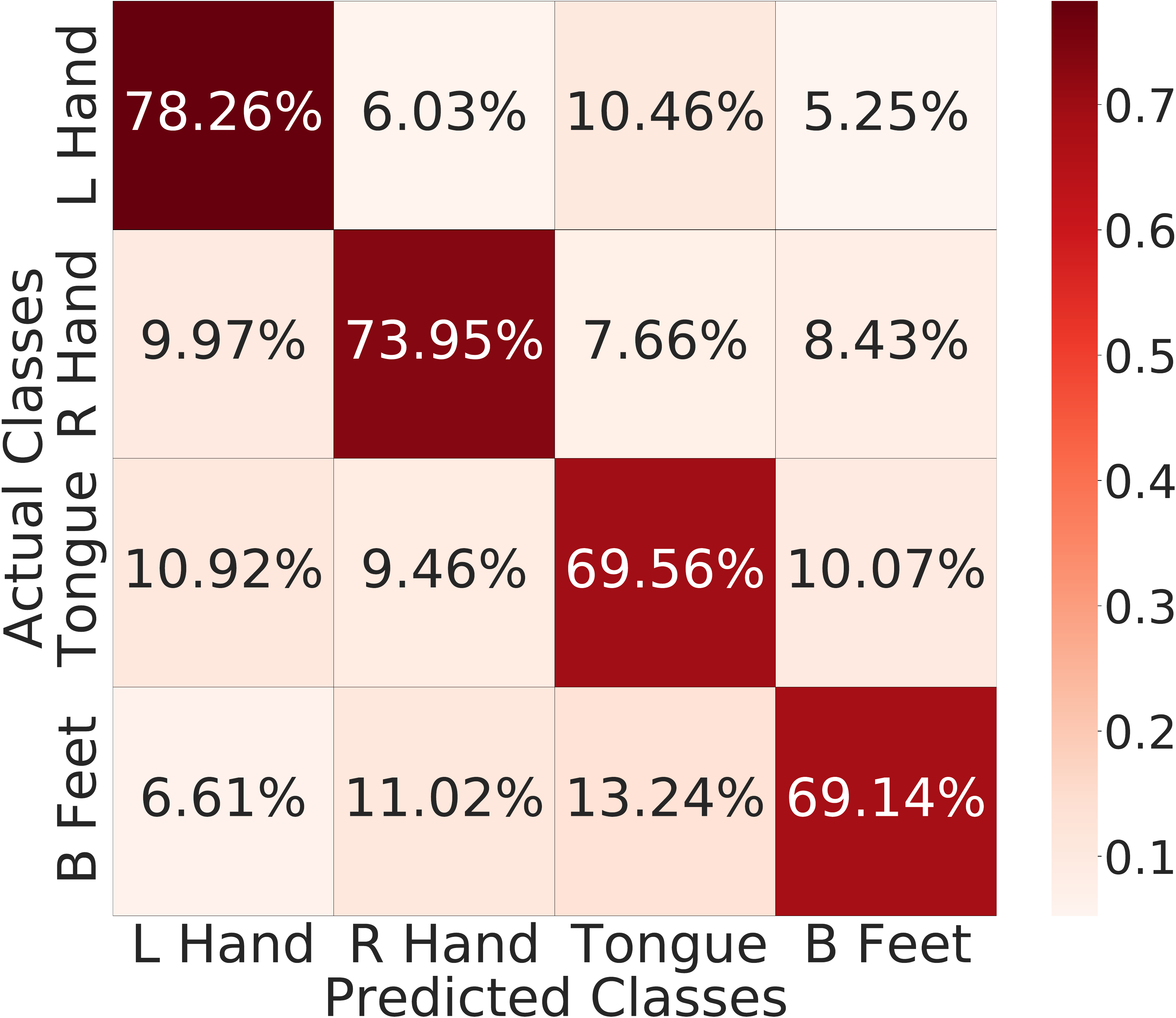}
            \caption[]%
            {{\small A05}}    
            \label{fig:cm_a05}
    \end{subfigure}%
    \\      
    % \hspace{-0.3cm}
    \begin{subfigure}[t]{0.2\textwidth}  
            \centering 
            \includegraphics[scale=0.052]{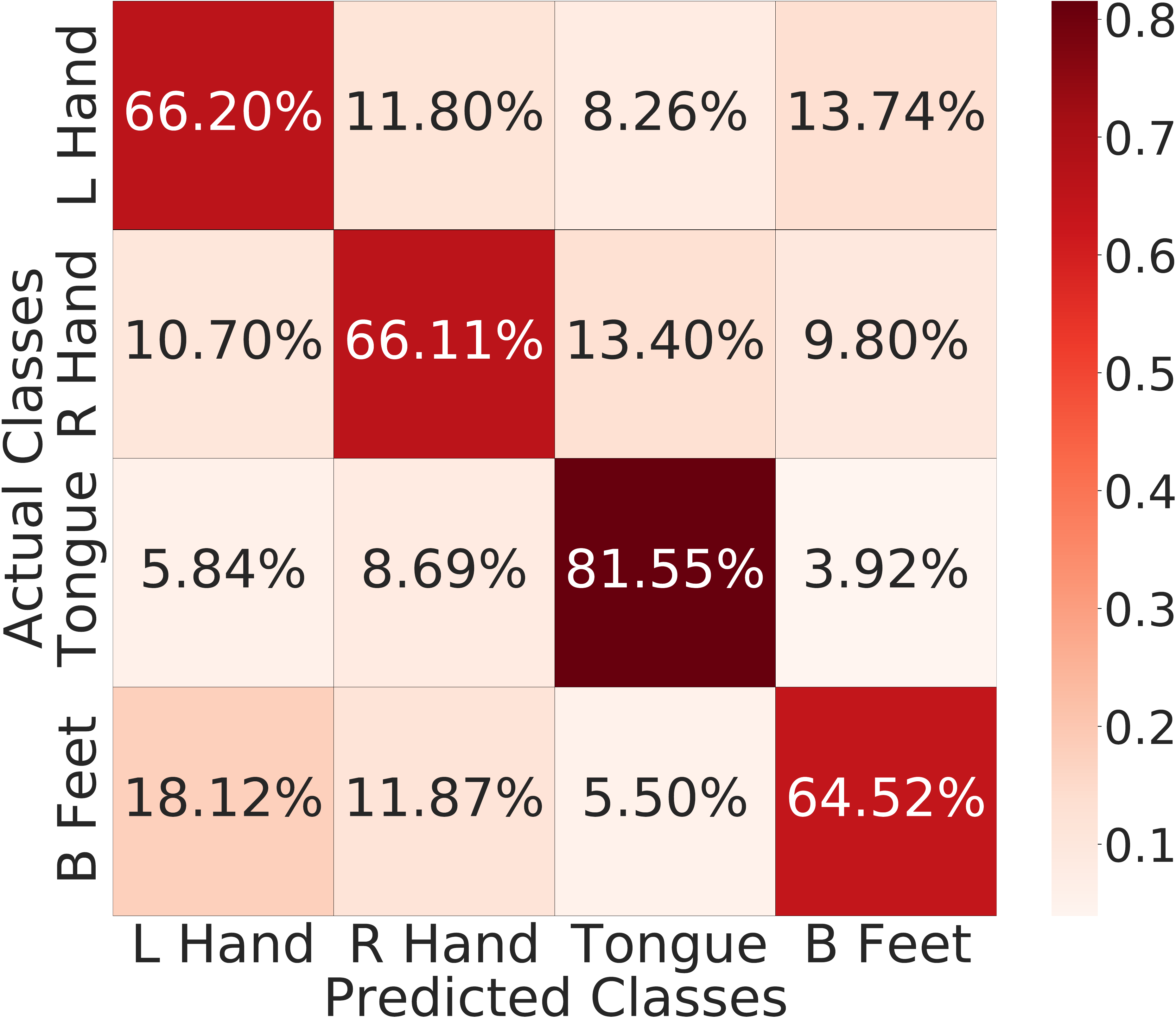}
            \caption[]%
            {{\small A06}}    
            \label{fig:cm_a06}
    \end{subfigure}%
    ~ 
    \hspace{-0.3cm}
    \begin{subfigure}[t]{0.2\textwidth}  
            \centering 
            \includegraphics[scale=0.052]{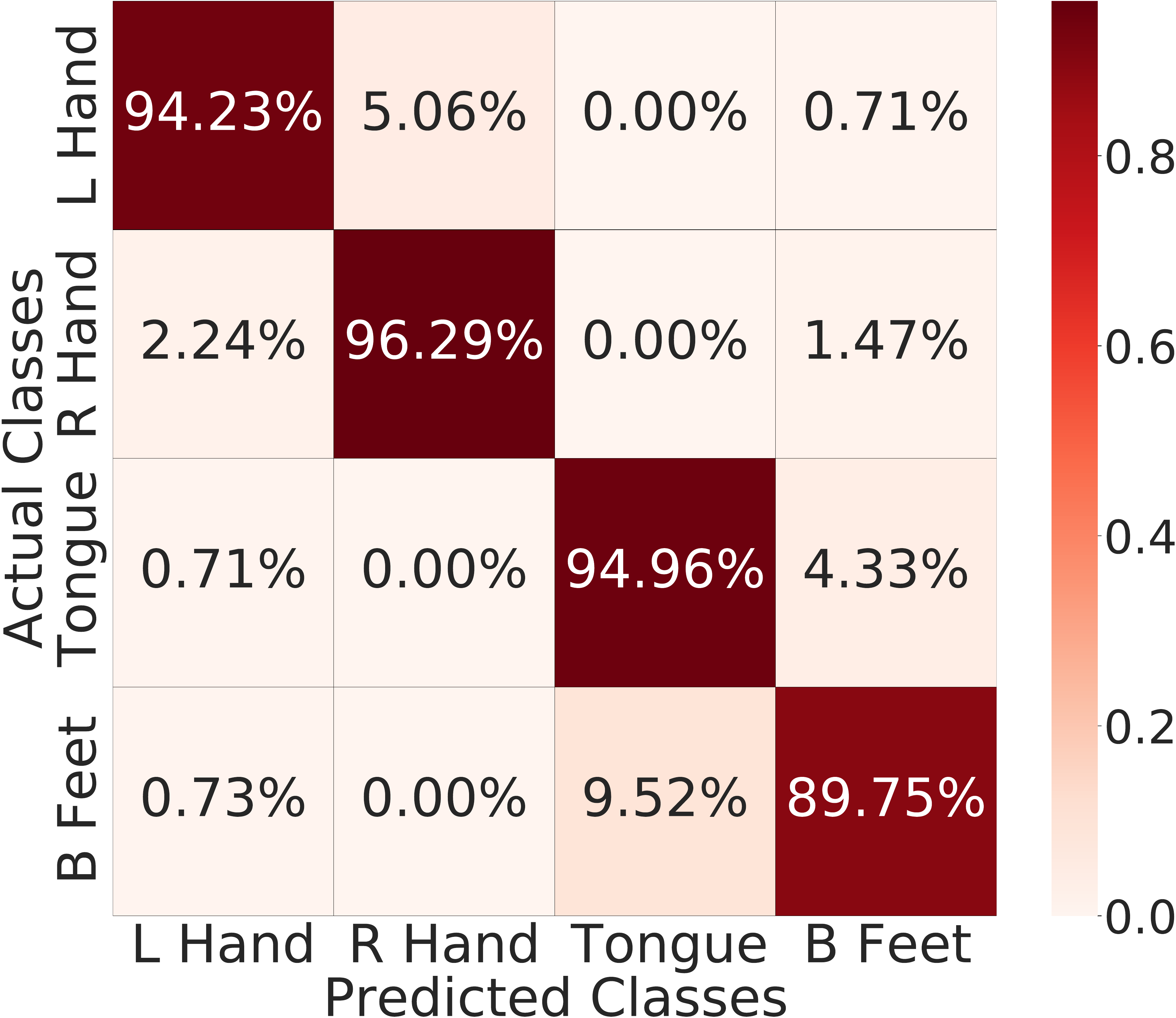}
            \caption[]%
            {{\small A07}}    
            \label{fig:cm_a07}
    \end{subfigure}%
    ~ 
        \hspace{-0.3cm}
        \begin{subfigure}[t]{0.2\textwidth}  
            \centering 
            \includegraphics[scale=0.052]{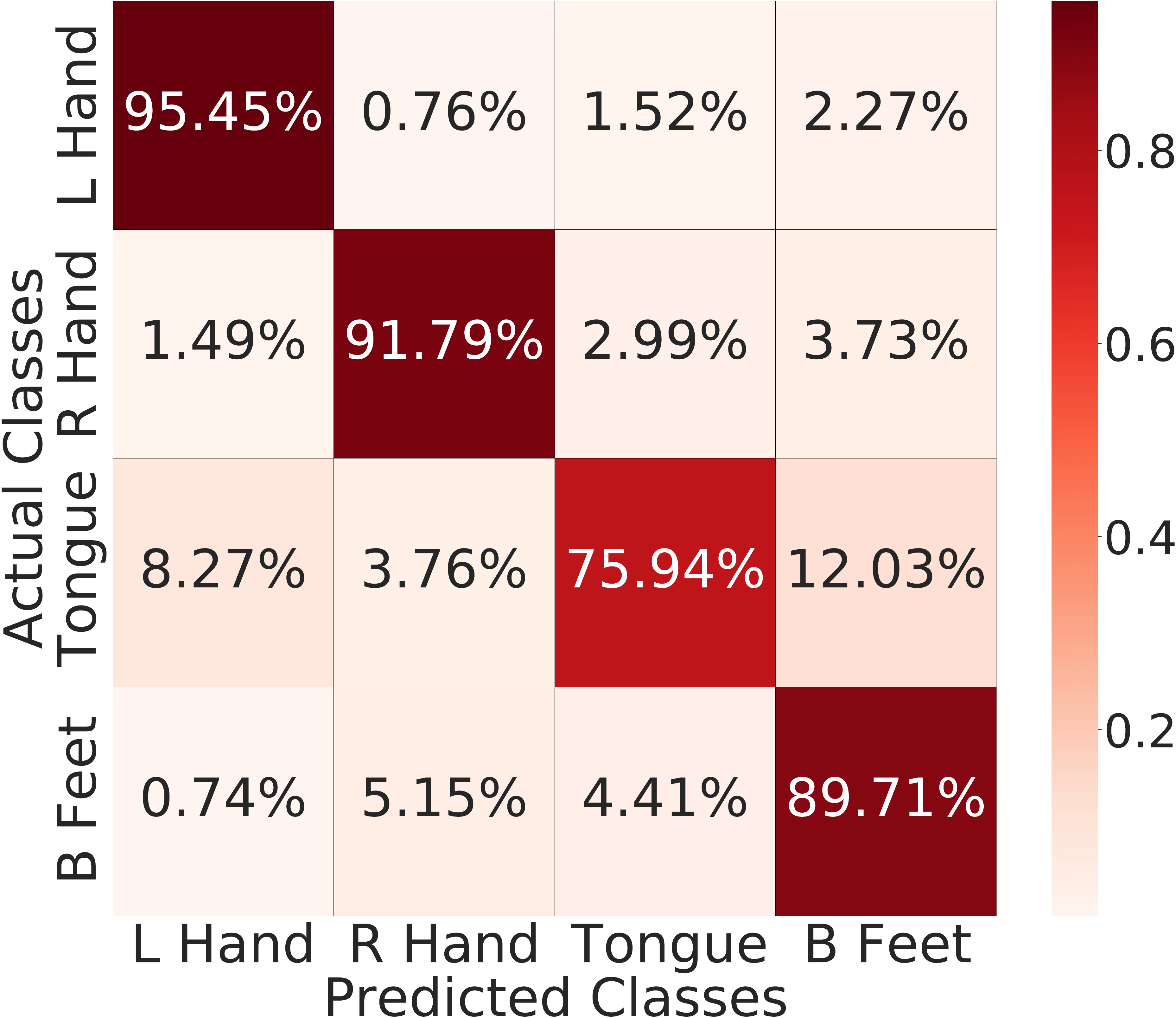}
            \caption[]%
            {{\small A08}}%PYD}}    
            \label{fig:cm_a08}
        \end{subfigure}%
    ~
        \hspace{-0.3cm}
        \begin{subfigure}[t]{0.2\textwidth}  
            \centering 
            \includegraphics[scale=0.052]{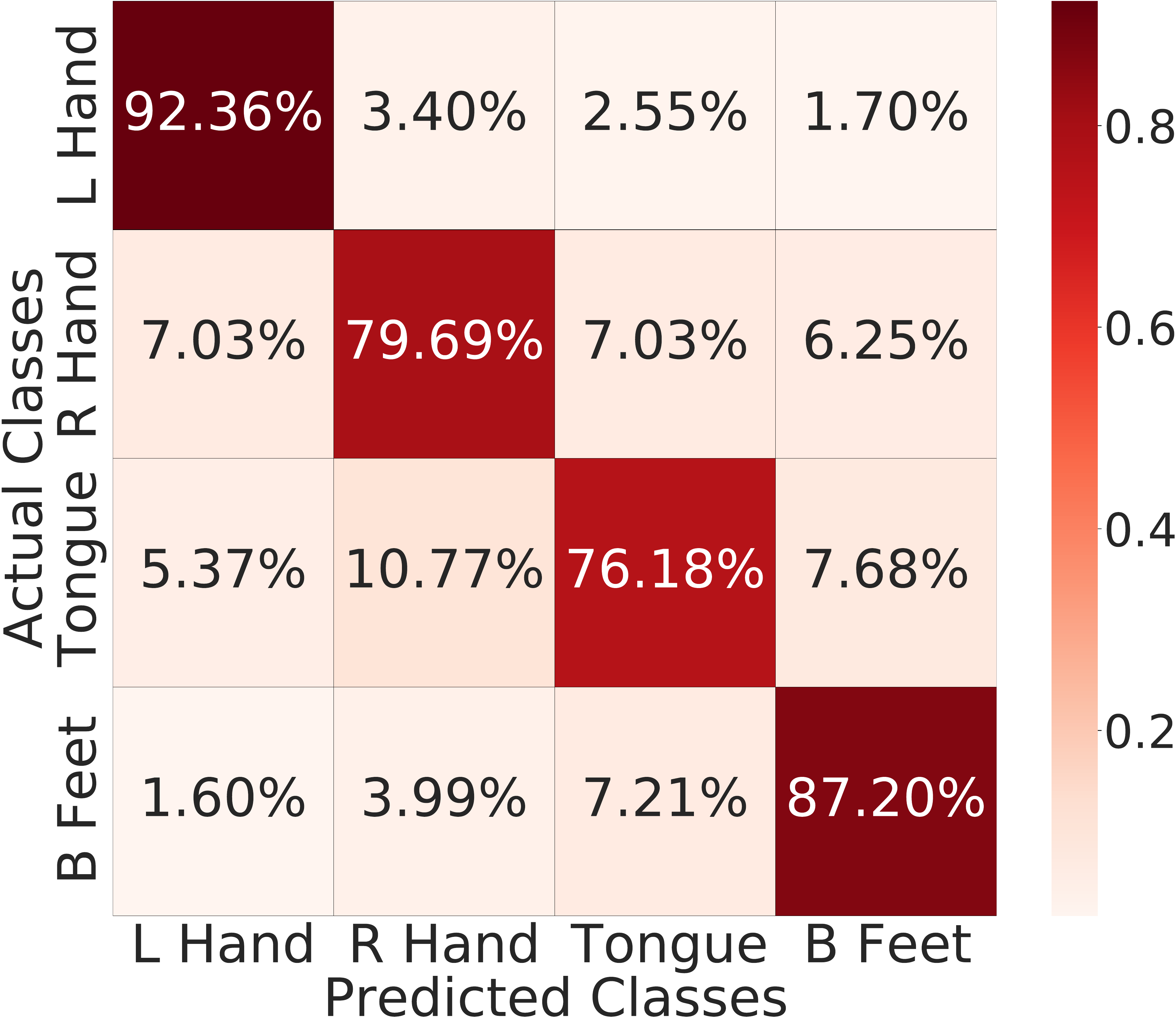}
            \caption[]%
            {{\small A09}}    
            \label{fig:cm_a09}
    \end{subfigure}%
    ~      
    \hspace{-0.3cm}
    \begin{subfigure}[t]{0.2\textwidth}  
            \centering 
            \includegraphics[scale=0.052]{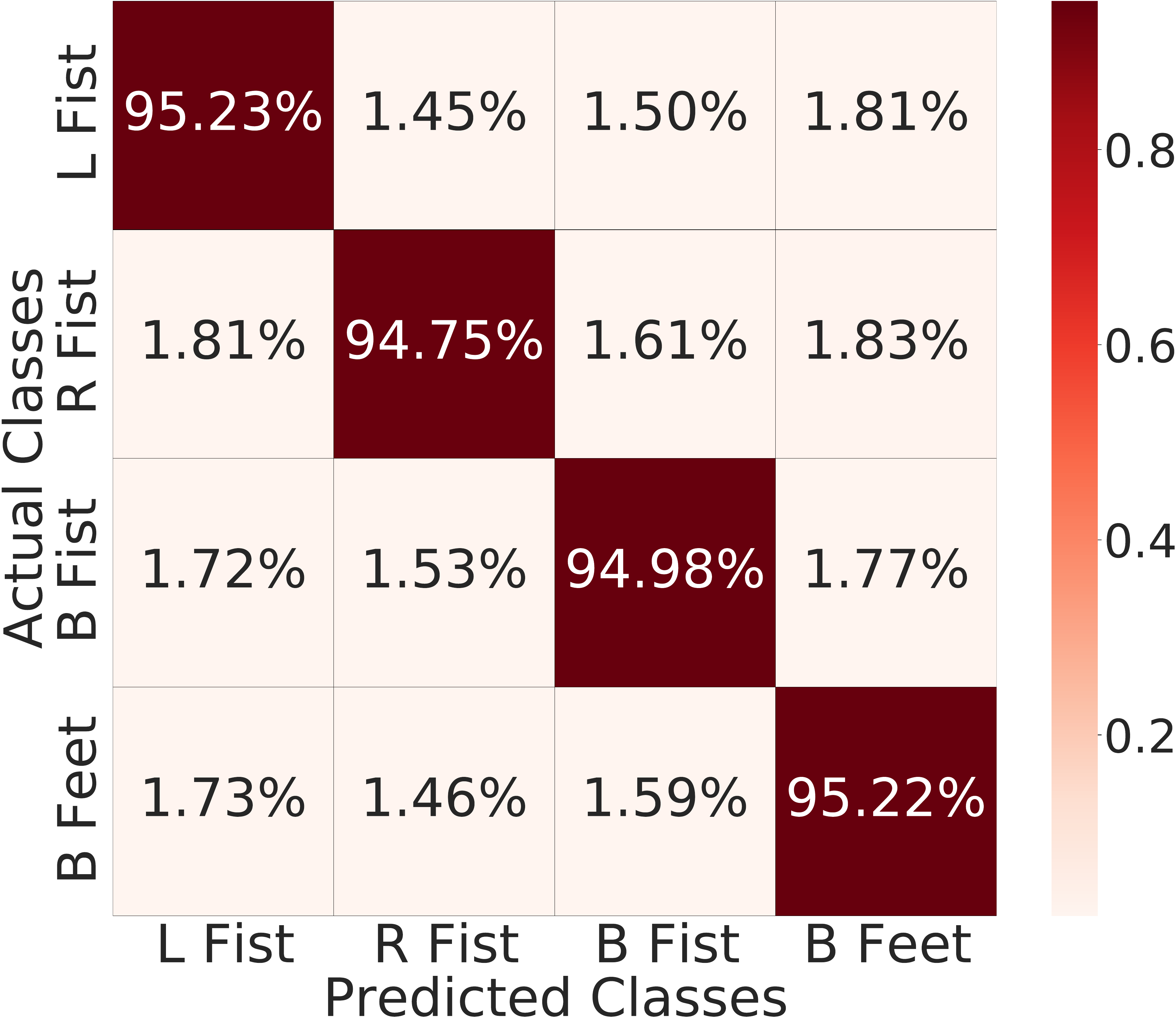}
            \caption[]%
            {{\small MI-PYD}}    
            \label{fig:cm_pyd}
    \end{subfigure}%

        \caption[  ]
        {\small Confusion matrices of the proposed method for MI-BCI and MI-PYD datasets.} 
        \label{fig:confusion_matrix}
\end{figure*}

\begin{figure*}[!th]
        \centering
        % \hspace{-0.3cm}
        \begin{subfigure}[t]{0.2\textwidth}  
            \centering 
            \includegraphics[scale=0.25]{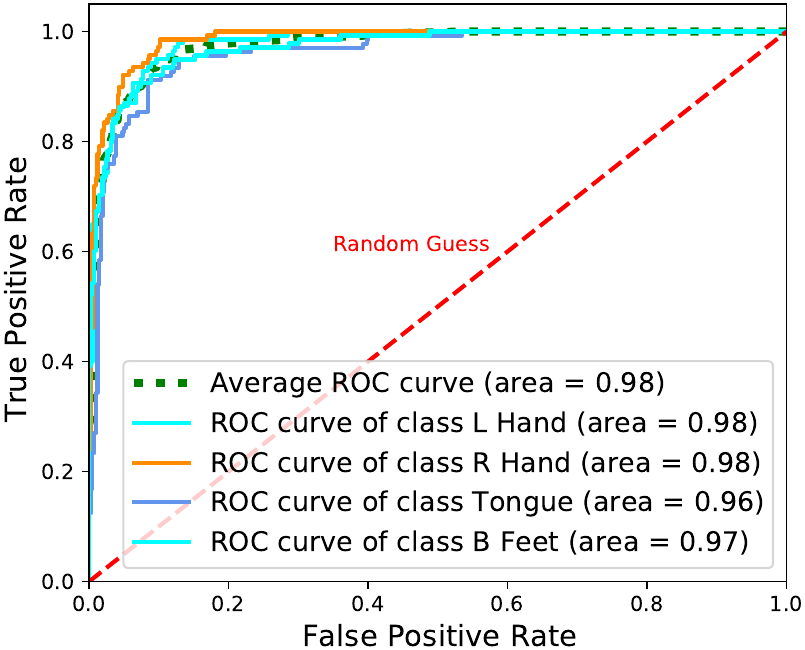}
            \caption[]%
            {{\small A01}}    
            \label{fig:roc_a01}
        \end{subfigure}%
    ~     
        \hspace{-0.3cm}
        \begin{subfigure}[t]{0.2\textwidth}  
            \centering 
            \includegraphics[scale=0.25]{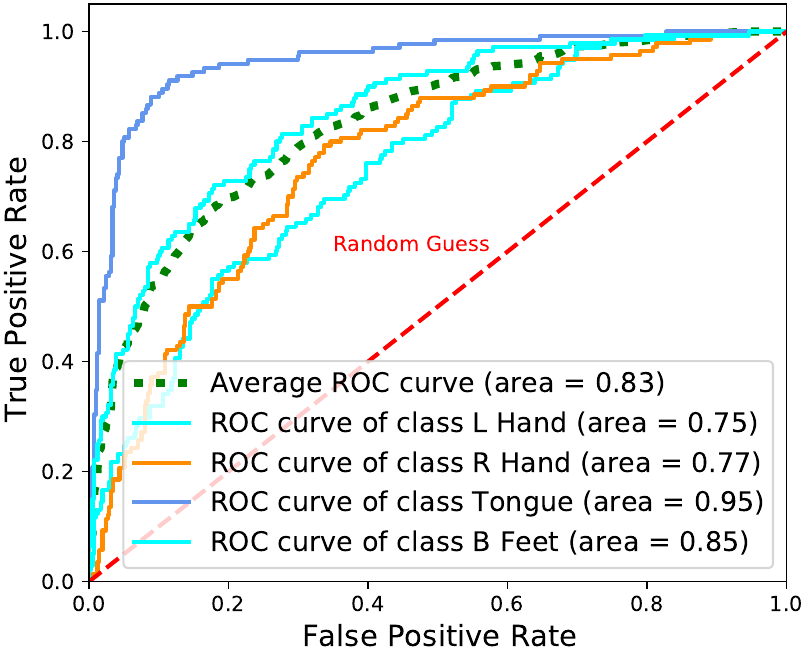}
            \caption[]%
            {{\small A02}}    
            \label{fig:roc_a02}
        \end{subfigure}%
    ~
        \hspace{-0.3cm}
        \begin{subfigure}[t]{0.2\textwidth}  
            \centering 
            \includegraphics[scale=0.25]{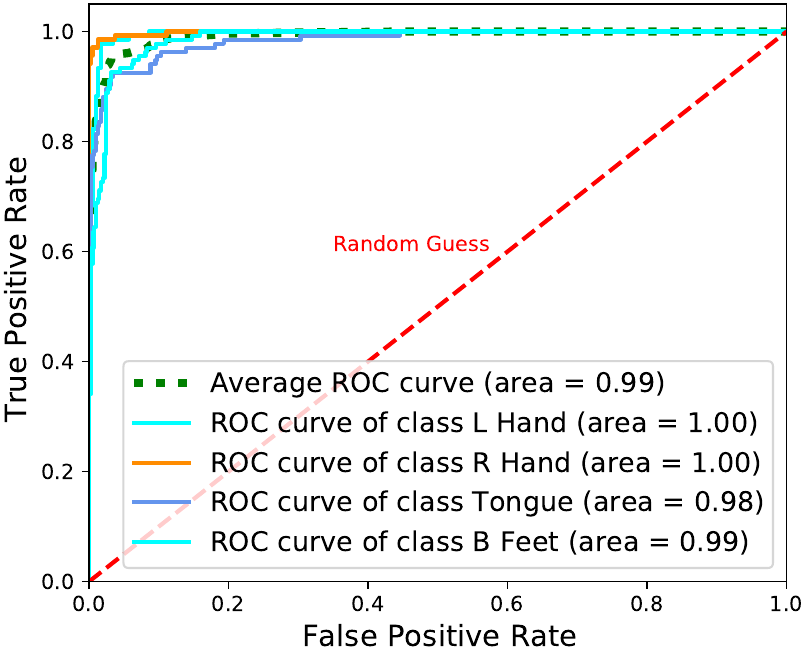}
            \caption[]%
            {{\small A03}}    
            \label{fig:roc_a03}
        \end{subfigure}%
    ~ 
        \hspace{-0.3cm}
        \begin{subfigure}[t]{0.2\textwidth}  
            \centering 
            \includegraphics[scale=0.25]{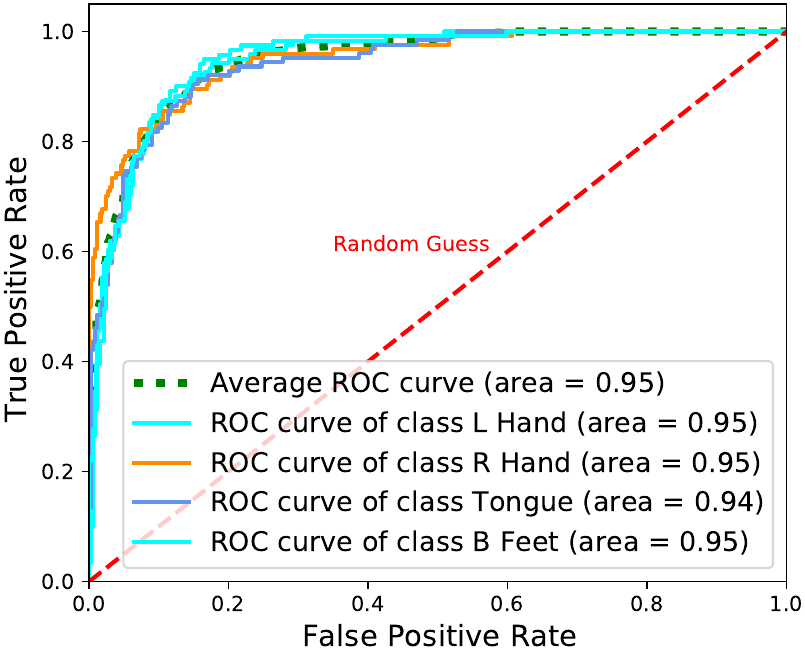}
            \caption[]%
            {{\small A04}}     
            \label{fig:roc_a04}
        \end{subfigure}%
    ~
        \hspace{-0.3cm}
        \begin{subfigure}[t]{0.2\textwidth}  
            \centering 
            \includegraphics[scale=0.25]{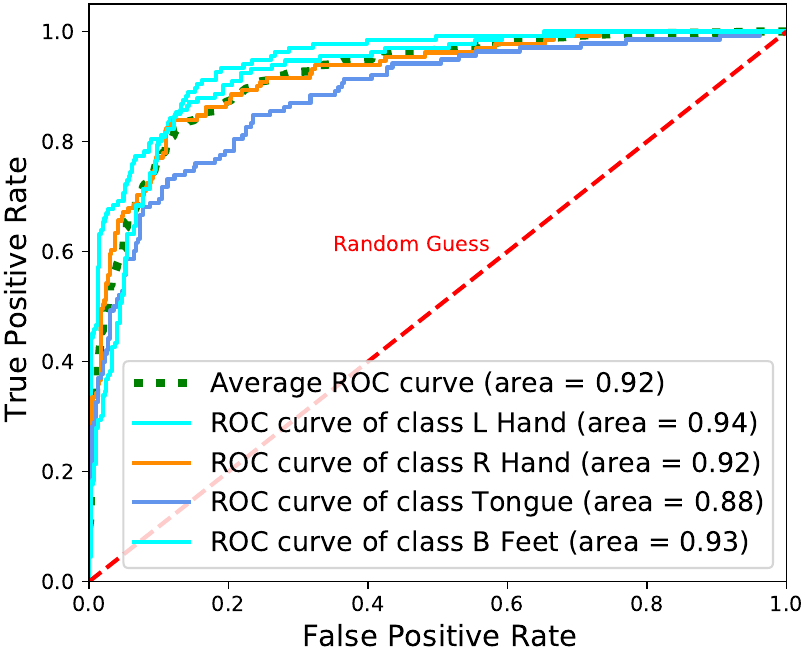}
            \caption[]%
            {{\small A05}}    
            \label{fig:roc_a05}
        \end{subfigure}%
    \\     
        % \hspace{-0.3cm}
        \begin{subfigure}[t]{0.2\textwidth}  
            \centering 
            \includegraphics[scale=0.25]{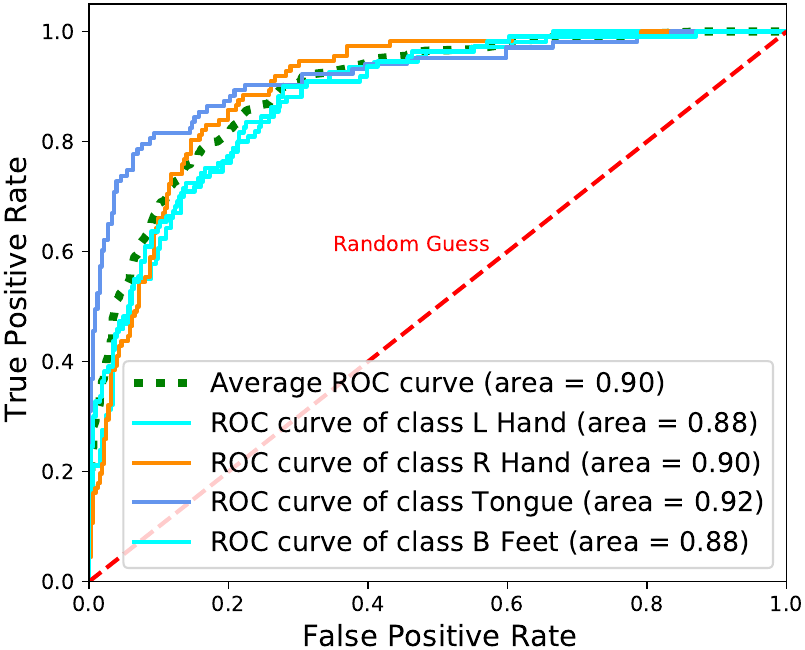}
            \caption[]%
            {{\small A06}}    
            \label{fig:roc_a06}
        \end{subfigure}%
    ~
        \hspace{-0.3cm}
        \begin{subfigure}[t]{0.2\textwidth}  
            \centering 
            \includegraphics[scale=0.25]{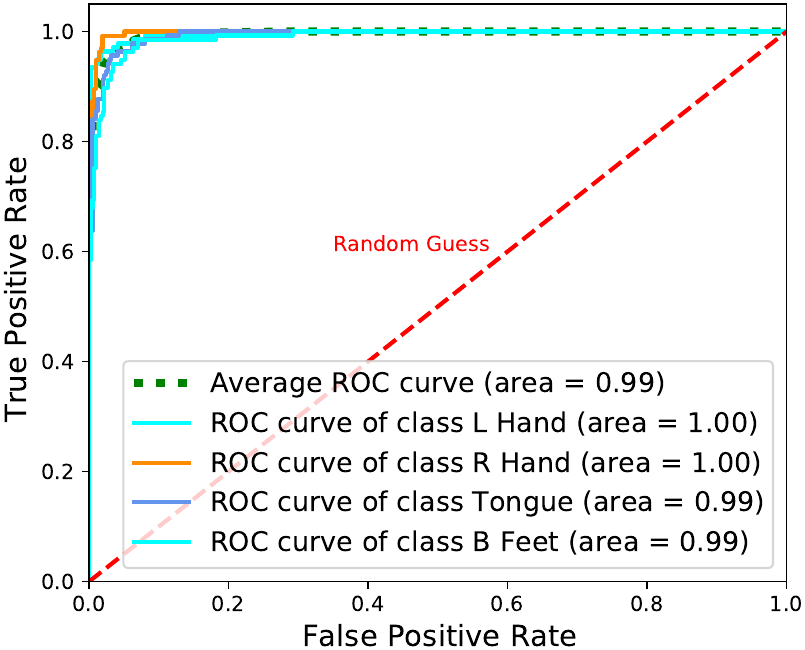}
            \caption[]%
            {{\small A07}}    
            \label{fig:roc_a07}
        \end{subfigure}%
    ~ 
        \hspace{-0.3cm}
        \begin{subfigure}[t]{0.2\textwidth}  
            \centering 
            \includegraphics[scale=0.25]{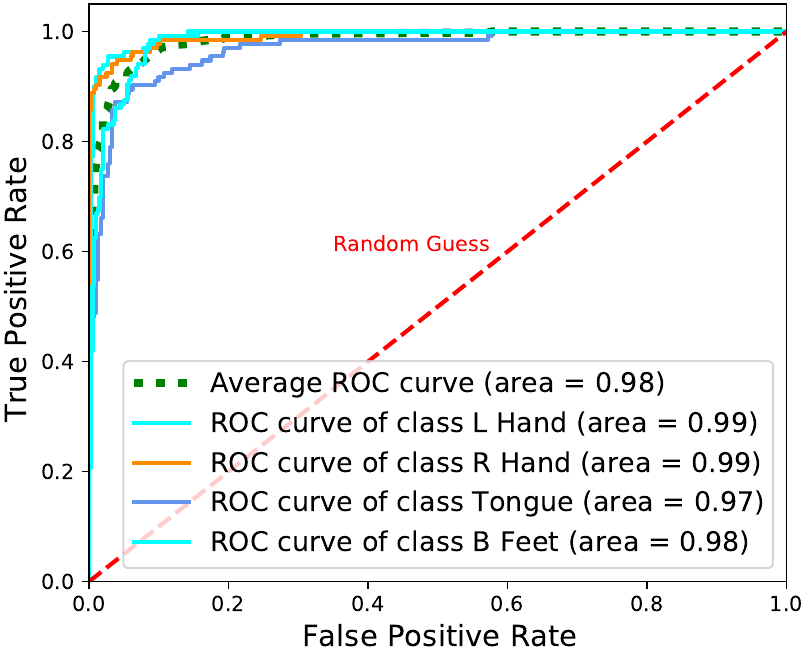}
            \caption[]%
            {{\small A08}}    
            \label{fig:roc_a08}
        \end{subfigure}%
    ~
        \hspace{-0.3cm}
        \begin{subfigure}[t]{0.2\textwidth}  
            \centering 
            \includegraphics[scale=0.25]{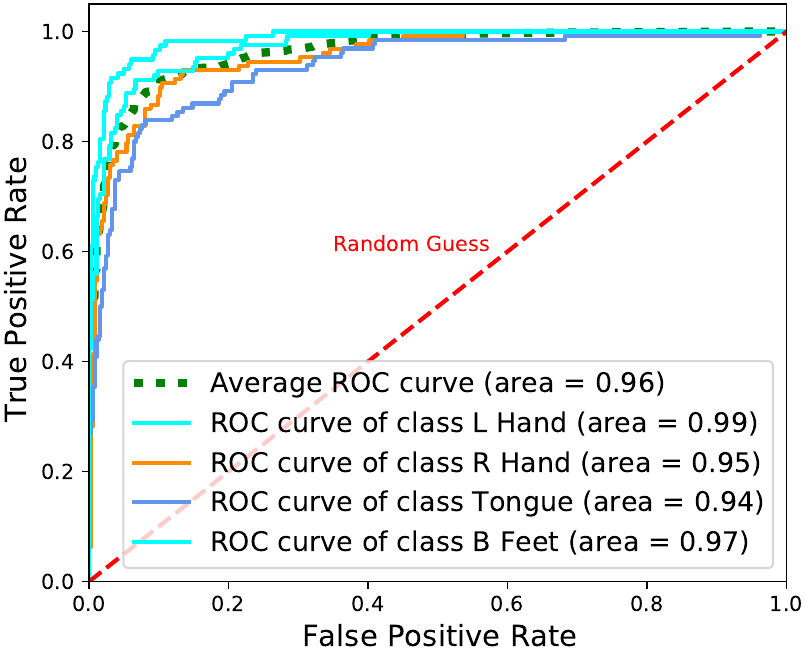}
            \caption[]%
            {{\small A09}}    
            \label{fig:roc_a09}
        \end{subfigure}%
    ~     
        \hspace{-0.3cm}
        \begin{subfigure}[t]{0.2\textwidth}  
            \centering 
            \includegraphics[scale=0.25]{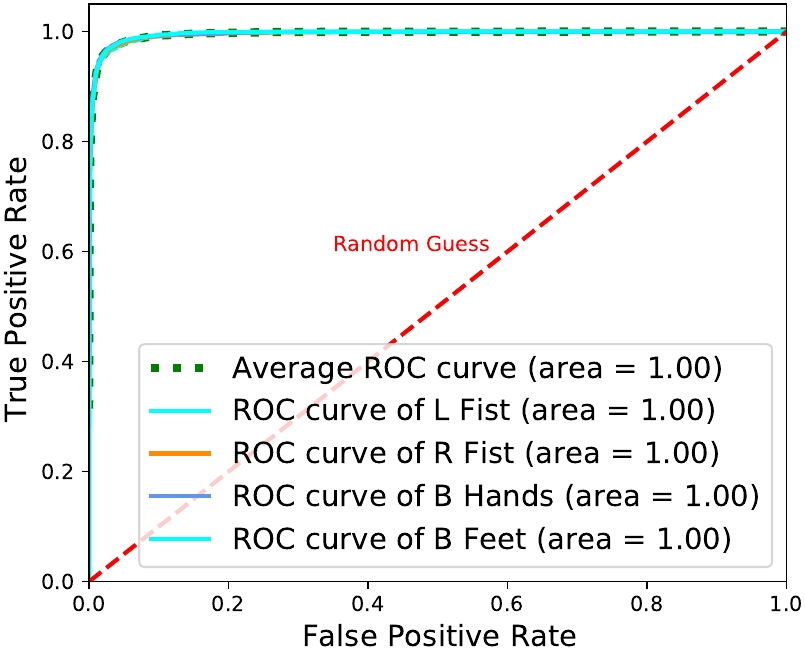}
            \caption[]%
            {{\small MI-PYD}}    
            \label{fig:roc_pyd}
        \end{subfigure}%
        \caption[  ]
        {\small ROC plots of the proposed method for MI-BCI and MI-PYD datasets.} 
        \label{fig:roc}
    \end{figure*}

Figure \ref{fig:confusion_matrix} shows the confusion matrices of both motor imagery datasets. In Figure \ref{fig:confusion_matrix}, the first nine confusion matrices corresponds to each subject of MI-BCI dataset and the last confusion matrix is of MI-PYD dataset. These figures show a balanced class-wise decoding performance of the proposed method for both datasets. While most of the confusion matrices show a balanced decoding performance, subject A02 of MI-BCI dataset shows a rather unbalanced confusion matrix. It shows that for subject A02, the decoding performance is high for only 'Tongue' task among all four task. This behaviour aligns with the related study \cite{musallam2021electroencephalography}. The reason behind this behaviour is the excessive noise and low signal-to-noise ratio in the data of subject A02. However, the performance is still better than the random probability. Moreover, the proposed method performed consistently in balanced decoding manner for the rest of the subjects of MI-BCI dataset.

For the analysis of the performance of the classifier, the computation of the true positive rate (sensitivity) and false-positive rate (1- sensitivity) at the various threshold of the classification confidence is performed to evaluate the classifier performance. The classifier with a higher true positive rate at a lower false-positive rate is considered better than others. For the sensitivity-specificity tradeoff analysis, a ROC plots has been drawn for each classification category for both analyzed datasets and shown in Figure \ref{fig:roc}. Figure \ref{fig:roc} shows ten ROC plots that includes one ROC plots for each subject of MI-BCI dataset and one ROC plots for MI-PYD dataset. A similar trend to the confusion matrices is seen in all ROC plots. All the ROC plots are showing balanced decoding across all the four MI tasks with an exception of subject A02 of MI-BCI dataset. This observation is consistent with the confusion matrix of the subject A02 of MI-BCI dataset. In the rest of the ROC plots, the decoding performance of each class is balanced. Also, the ROC plot of MI-PYD dataset shows an almost perfect decoding performance. Therefore, the good performance of the proposed method for both datasets validates the need for feature reweighting in MI-EEG classification.

\begin{table}[!t]
\centering
\small

\caption{Comparison of the proposed method with various state-of-the-art (SOTA) methods for four-class motor imagery EEG signal classification of MI-BCI dataset. Best values are bold-faced.}

\begin{tabular}{l|ll}
\hline

\textbf{Method} &   \textbf{Accuracy} &  \textbf{F-Measure}   \\ \hline

FBCSP \cite{ang2008filter}   & 69.56 $\pm$ 13.58 & 0.69 $\pm$ 0.14 \\
DeepConvNet \cite{schirrmeister2017deep}   & 65.66 $\pm$ 15.90 & 0.64 $\pm$ 0.17 \\
ShallowConvNet \cite{schirrmeister2017deep}   & 76.85 $\pm$ 13.66 & 0.77 $\pm$ 0.14 \\
EEGNet \cite{lawhern2018eegnet}    & 62.39 $\pm$ 9.32 & 0.62 $\pm$ 0.09 \\
EEGNet Fusion \cite{roots2020fusion} & 66.39 $\pm$ 8.91 & 0.62 $\pm$ 0.09 \\
MI-EEGNet \cite{riyad2021mi}    & 75.85 $\pm$ 11.93 & 0.76 $\pm$ 0.12 \\
TS-SEFFNet \cite{li2021temporal}    & 53.43 $\pm$ 12.49 & 0.50 $\pm$ 0.14 \\
LMDA  \cite{miao2023lmda}       & 55.91 $\pm$ 12.98 & 0.56	$\pm$ 0.13 \\
Proposed Method       & \textbf{80.67} $\pm$ 11.30  & \textbf{0.80} $\pm$ 0.11 \\ \hline

\end{tabular}
\label{table:comparison_results_bci}
\end{table}

\subsection{Comparative Analysis}

% Along with the performance of these datasets, a rigorous comparison has also been reported. 
The methods considered for the comparison are filter bank common spatial patterns (FBCSP) \cite{ang2008filter}, DeepConvNet \cite{schirrmeister2017deep}, ShallowConvNet \cite{schirrmeister2017deep}, EEGNet \cite{lawhern2018eegnet}, EEGNet Fusion  \cite{roots2020fusion}, MI-EEGNet \cite{riyad2021mi}, TS-SEFFNet  \cite{li2021temporal} and LMDA \cite{miao2023lmda}. As it is prominent that most of the BCI studies have used different experimental settings and performance comparison of methods with different experimental settings can be misleading. Therefore, for fair comparison, all of the above mentioned compared methods are trained and evaluated with same experimental settings as the proposed method.

The EEGNet \cite{lawhern2018eegnet} is a sequential model with a stack of convolutional kernels and lacks the effective use of different kernel sizes for the same subject, limiting its performance. The same is reflected in Table \ref{table:comparison_results_bci} and \ref{table:comparison_results_pyd}. The EEGNet Fusion \cite{roots2020fusion} and MI-EEGNet \cite{riyad2021mi} are based on multi-branch architecture and are superior in nature over EEGNet \cite{lawhern2018eegnet}. The TS-SEFFNet \cite{li2021temporal} extracts distinct temporal and spectral features separately and fuses them using the squeeze and excitation approach, but extraction of efficient channel-wise features was not employed. Table \ref{table:comparison_results_bci} can infer that this method struggles to extract meaningful features as all the metrics are the lowest for both datasets. 

DeepConvNet and ShallowConvNet \cite{schirrmeister2017deep} are convolutional neural network-based methods performing better than many compared methods. DeepConvNet has a deep neural architecture and can learn highly variable MI-EEG signals. But it over-fits upon training, and the performance degrades because of irrelevant features in MI-EEG signals. In contrast, ShallowConvNet is a comparatively shallower network that helps to decode discriminative features and does not overfit. A shallower network can work better for the data with fewer samples. However, a shallower network will underperform when data is vast, as it cannot extract meaningful information from such a large and variable dataset. Therefore, from Table \ref{table:comparison_results_bci}, it can be observed that the ShallowConvNet method outperforms all the other compared methods for the MI-BCI dataset, as the number of samples is smaller, and other methods overfit very frequently. Our proposed method is also a deep CNN-based method. Still, it can produce much better results than all of the compared methods as it can deduce the most contributing features and ignores the less contributing features from the data. It can be confirmed from Table \ref{table:comparison_results_bci}. 

LMDA \cite{miao2023lmda} is a shallow CNN-based method that utilizes channel and depth attention techniques. While this method is lightweight, it struggles to extract relevant temporal information from MI-EEG signals. Also, the MI-EEG data have low signal-to-noise ratio and high amount of irrelevant information. The results in Table \ref{table:comparison_results_bci} validates the inefficient classification performance of LMDA method for both the datasets. Among all of the compared methods, MI-EEGNet \cite{riyad2021mi} performed good for both datasets. The proposed architecture has achieved  $4.82\%$ and $9.34\%$ higher classification accuracy than MI-EEGNet method for MI-BCI and MI-PYD dataset, respectively.

\begin{table}[!t]
\centering
\small
% \begin{small}

\caption{Comparison of the proposed method with various state-of-the-art (SOTA) methods for four-class motor imagery EEG signal classification of MI-PYD dataset. Best values are bold-faced.}

\begin{tabular}{l|ll}
\hline

\textbf{Method} &   \textbf{Accuracy} &  \textbf{F-Measure}   \\ \hline

FBCSP \cite{ang2008filter}         & 30.63 $\pm$ 0.53  & 0.30 $\pm$ 0.01 \\
DeepConvNet \cite{schirrmeister2017deep}        & 57.66 $\pm$ 2.23  & 0.57 $\pm$ 0.03 \\
ShallowConvNet \cite{schirrmeister2017deep}      &  70.16 $\pm$ 0.28  & 0.70 $\pm$ 0.00 \\
EEGNet \cite{lawhern2018eegnet}         & 37.53 $\pm$ 1.63  & 0.36 $\pm$ 0.03 \\
EEGNet Fusion \cite{roots2020fusion}  & 73.73 $\pm$ 1.32  & 0.74 $\pm$ 0.01 \\
MI-EEGNet \cite{riyad2021mi}      & 85.71 $\pm$ 3.52  & 0.86 $\pm$ 0.04 \\
TS-SEFFNet \cite{li2021temporal}     & 57.31 $\pm$ 1.06  & 0.57 $\pm$ 0.01 \\
LMDA  \cite{miao2023lmda}          & 32.70 $\pm$ 0.66 &  0.32 $\pm$ 0.01 \\
Proposed Method       & \textbf{95.05} $\pm$ 2.11  & \textbf{0.95} $\pm$ 0.02 \\ \hline

\end{tabular}
\label{table:comparison_results_pyd}
% \end{footnotesize}
\end{table}

Additionally, the famous traditional machine learning method FBCSP \cite{ang2008filter} is also considered for comparison. The proposed method performed better than the FBCSP in both MI-BCI and MI-PYD dataset. However, Table \ref{table:comparison_results_bci} shows that for the MI-BCI dataset, FBCSP has performed superior to four of the other CNN-based methods (EEGNet, EEGNet Fusion, DeepConvNet, LMDA). It shows the significance of extracting features using filter bank and common spatial patterns over conventional CNN-based approaches such as EEGNet, EEGNet Fusion, DeepConvNet and LMDA method. However, the usage of feature reweighting is found to be superior as the feature reweighting mechanism helps to filter out irrelevant information efficiently. As a result, the proposed method outperforms FBCSP. Thus, proposed method performed significantly better than all of the compared methods for MI-BCI dataset.

\begin{figure}[!t]
\centering
 \includegraphics[width=\linewidth]{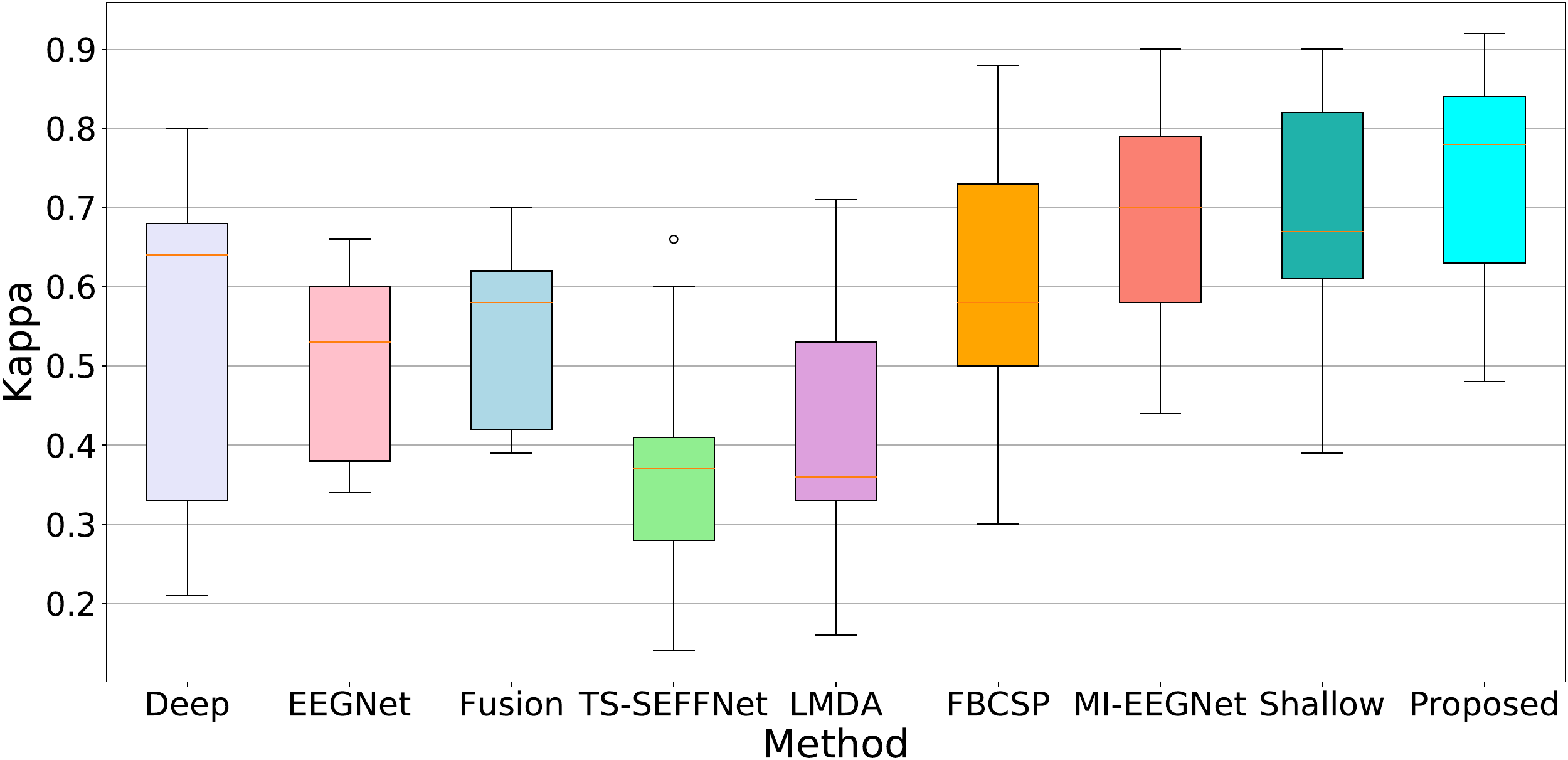}
  \caption{\textcolor{black}{Kappa measure comparison of proposed method and compared methods for first motor imagery dataset (MI-BCI dataset).} }
  \label{fig:kappa_bci}
\end{figure}

\begin{figure}[!t]
\centering
 \includegraphics[width=\linewidth]{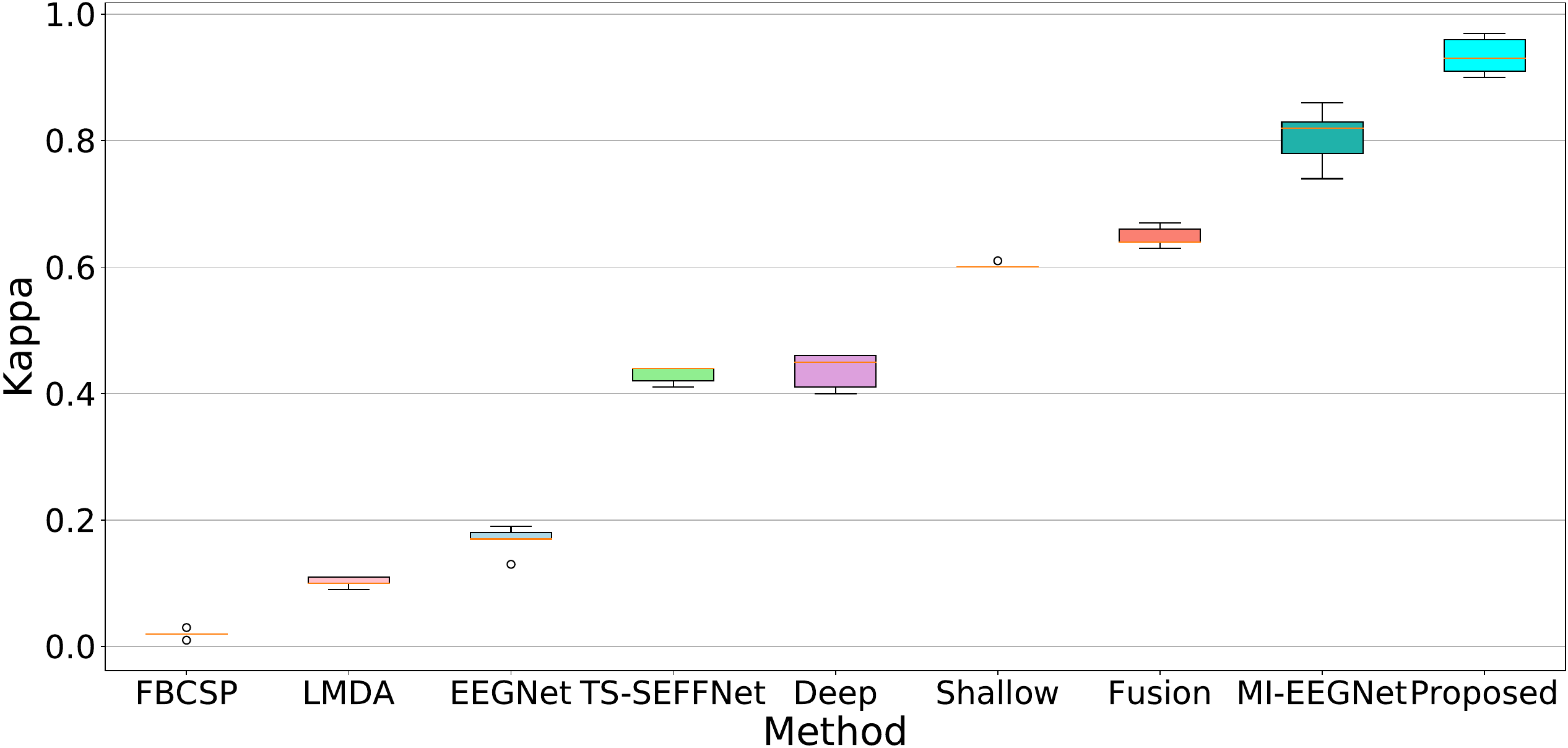}
  \caption{\textcolor{black}{Kappa measure comparison of proposed method and compared methods for second motor imagery dataset (MI-PYD dataset).} }
  \label{fig:kappa_pyd}
\end{figure}

\textcolor{black}{ Figure \ref{fig:kappa_bci} shows the comparison of kappa values of proposed and compared methods for MI-BCI dataset. This figure shows that the mean kappa value of the proposed method is better than all compared methods. Moreover, the box range of the proposed method is also better than that of the compared methods. It indicates that proposed method has performed better than the compared methods over all folds of the 5-fold cross-validation. Similarly, Figure \ref{fig:kappa_pyd} depicts the kappa value comparison of the proposed method with the compared methods for MI-PYD dataset. This plot clearly indicates the superior classification performance of the proposed method for MI-PYD dataset. These kappa measure plots suggest a higher classification performance ability of the proposed method, which further validates the necessity of feature reweighting.}

\begin{table}[!t]
\centering
\small
% \begin{small}

\caption{Statistical significance p-values of paired t-test with different compared methods.}

\begin{tabular}{l|cc}
\hline
 \multicolumn{1}{c|}{\multirow{2}{*}{\textbf{Method}} } &  \multicolumn{2}{c}{\textbf{p-value}}  \\ \cline{2-3} 
 \multicolumn{1}{c|}{}   
  % \rule{0pt}{1\normalbaselineskip} 
  & \textbf{MI-BCI} & \textbf{MI-PYD} \\  \hline
FBCSP \cite{ang2008filter}  & $<0.001$    &   $<0.001$  \\
DeepConvNet \cite{schirrmeister2017deep}    & $<0.001$    &  $<0.001$   \\
ShallowConvNet \cite{schirrmeister2017deep}  & $0.059$    &  $<0.001$   \\
EEGNet \cite{lawhern2018eegnet} & $<0.001$    &  $<0.001$   \\
EEGNet Fusion \cite{roots2020fusion} & $<0.001$    &  $<0.001$   \\
MI-EEGNet \cite{riyad2021mi} &  $<0.001$   & $0.005$    \\
TS-SEFFNet \cite{li2021temporal}  &  $<0.001$   &  $<0.001$   \\
LMDA  \cite{miao2023lmda} &   $<0.001$  &  $<0.001$   \\  \hline
\end{tabular}
\label{table:significance}
% \end{footnotesize}
\end{table}

For the MI-BCI dataset, the proposed method shows better classification accuracy than the compared methods with a margin of $3.82\%$ in accuracy. A similar trend can be observed for the F-measure. The p-value of the proposed method versus all compared methods is less than $0.05$ as shown in Table \ref{table:significance}. The p-value for ShallowConvNet method is $0.059$ and this value is pretty close to the permissible p-value of less than $0.05$. Therefore, it can be concluded that statistically proposed method is marginally less significant than ShallowConvNet method. However, considering the $3.82\%$ accuracy difference between ShallowConvNet and the proposed method, it is evident that proposed method outperformed ShallowConvNet method. Table \ref{table:comparison_results_pyd} shows that for the MI-PYD dataset, the proposed method outperforms all the compared methods by a considerable margin in terms of accuracy and f-measure. The statistical significance values depicted in Table \ref{table:significance} confirms that the proposed method is better than all compared methods in both the datasets.

\subsection{\textcolor{black}{Performance on speech imagery task}}

\begin{table}[!t]
\centering
% \normalsize
% \begin{scriptsize}

\caption{\textcolor{black}{Comparison of the proposed method with various state-of-the-art (SOTA) methods for speech imagery tasks of SI-ASU dataset. Best values are bold faced.}}

\begin{tabular}{l|ll}
\hline

\textbf{Method} &   \textbf{Accuracy} &  \textbf{F-Measure}   \\ \hline

FBCSP \cite{ang2008filter}  & 54.15 $\pm$ 6.68 & 0.54 $\pm$ 0.07   \\
DeepConvNet \cite{schirrmeister2017deep}       & 39.61 $\pm$ 3.27 & 0.37 $\pm$ 0.04  \\ 	
ShallowConvNet \cite{schirrmeister2017deep}    & 68.89 $\pm$ 4.11 & 0.69 $\pm$ 0.04  \\ 	
EEGNet \cite{lawhern2018eegnet}         & 38.89 $\pm$ 3.01 & 0.39 $\pm$ 0.03  \\ 	
EEGNet Fusion \cite{roots2020fusion}   & 39.30 $\pm$ 4.08 & 0.39 $\pm$ 0.04  \\ 	
MI-EEGNet \cite{riyad2021mi}        & 68.04 $\pm$ 2.49 & 0.68 $\pm$ 0.03  \\ 	
TS-SEFFNet \cite{li2021temporal}        & 50.22 $\pm$ 3.27 & 0.50 $\pm$ 0.03  \\ 	
LMDA  \cite{miao2023lmda}           & 43.74 $\pm$ 5.60 & 0.43 $\pm$ 0.06  \\ 	
Proposed Method         & \textbf{74.13 $\pm$ 2.26}  & \textbf{0.74 $\pm$ 0.02} \\ \hline
\end{tabular}
\label{tab:sota_asu}
% \end{scriptsize}
\end{table}

\textcolor{black}{The generalizability of the proposed method to tasks beyond motor imagery was evaluated using SI-ASU dataset, a speech imagery dataset. Speech imagery involves the mental simulation of speech, where participants imagine the articulation of phonemes or words without vocalizing them. SI-ASU dataset contains EEG recordings of participants imagining the speech of short words ('in,' 'out,' and 'up'). Decoding speech imagery holds significant importance for individuals with speech disabilities, such as patients with locked-in syndrome or those recovering from stroke \cite{panachakel2021decoding}. Additionally, speech imagery has potential applications in gaming, neuro-marketing, and robotic arm control \cite{zhang2024speech}.}

\begin{figure}[!t]
\centering
 \includegraphics[width=\linewidth]{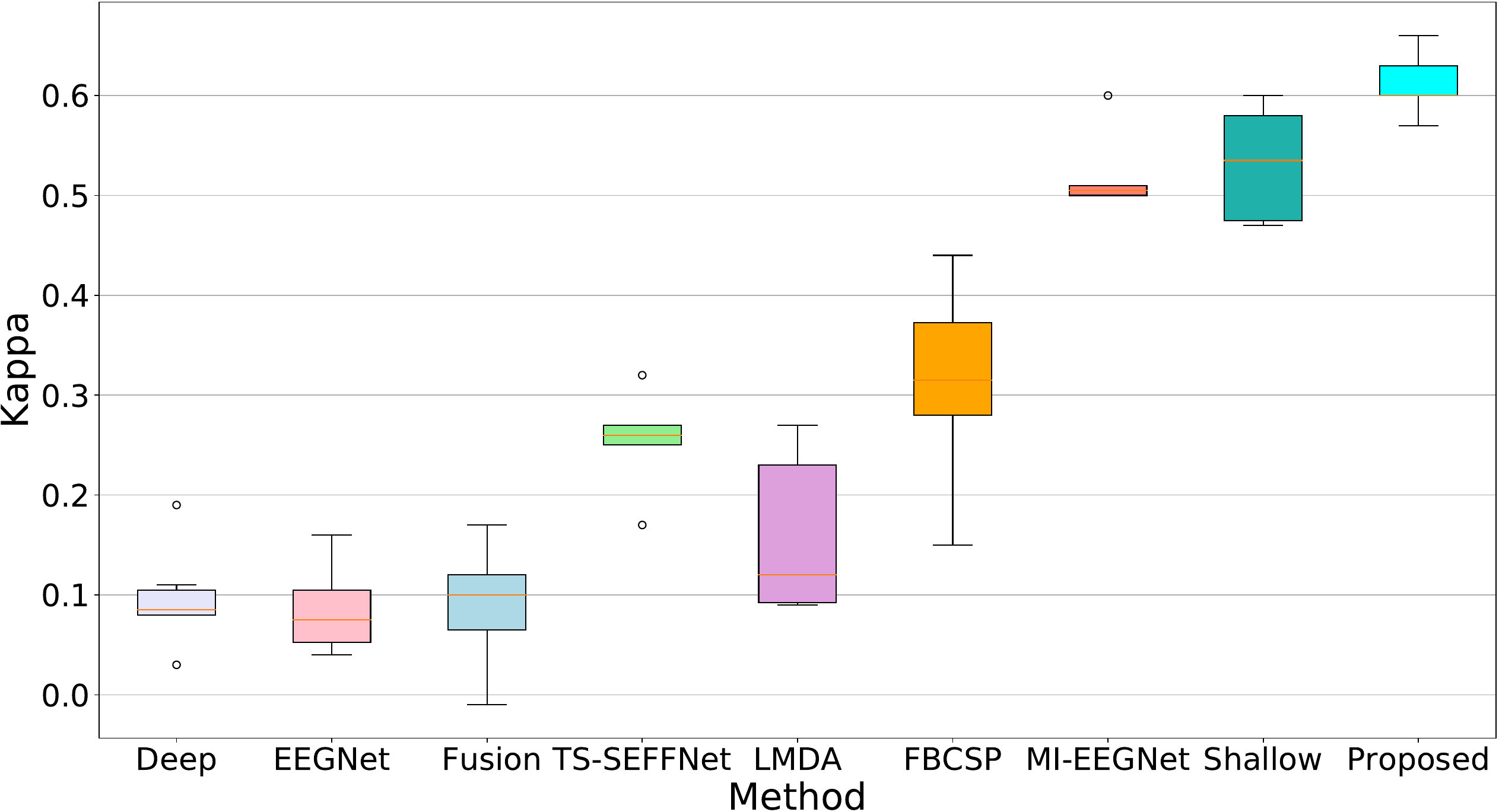}
  \caption{\textcolor{black}{Kappa measure comparison of proposed method and compared methods for speech imagery dataset (SI-ASU dataset).} }
  \label{fig:kappa_asu}
\end{figure}

\textcolor{black}{The accuracy of the proposed method and the compared methods on SI-ASU dataset is summarized in Table \ref{tab:sota_asu}. The accuracy and f-measure values in Table \ref{tab:sota_asu} indicate that the proposed method has outperformed the compared methods by a huge margin. It indicates the better generalizability and robustness of the proposed method on a task different from motor imagery. Moreover, the kappa value comparison depicted in Figure \ref{fig:kappa_asu} shows that the class separability of the proposed method is better than all of the compared methods. The median of the kappa value of the proposed method is more than that of all compared methods, and the variance is also less. It suggests that the proposed method has performed exceptionally well for the speech imagery task EEG dataset.}

\subsection{Performance on motor movement task}

\begin{table}[!t]
\centering
% \normalsize
% \begin{scriptsize}

\caption{Comparison of the proposed method with various state-of-the-art (SOTA) methods for motor movement tasks of MM-HGD dataset. Best values are bold faced and second best are underlined.}

\begin{tabular}{l|ll}
\hline

\textbf{Method} &   \textbf{Accuracy} &  \textbf{F-Measure}   \\ \hline

FBCSP \cite{ang2008filter}  & 74.40 $\pm$ 8.80 & 0.74 $\pm$ 0.09  \\
DeepConvNet \cite{schirrmeister2017deep}           & 84.43 $\pm$ 11.04 & 0.84 $\pm$ 0.13 \\ 	
ShallowConvNet \cite{schirrmeister2017deep}   & \textbf{92.93} $\pm$ 4.38 & \textbf{0.93} $\pm$ 0.05 \\ 
EEGNet \cite{lawhern2018eegnet}          & 84.04 $\pm$ 4.04 & 0.83 $\pm$ 0.04 \\ 	
EEGNet Fusion \cite{roots2020fusion}   & 84.70 $\pm$ 3.79 & 0.84 $\pm$ 0.04 \\ 	
MI-EEGNet \cite{riyad2021mi}        & 89.36 $\pm$ 6.36 & 0.90 $\pm$ 0.06 \\ 	
TS-SEFFNet \cite{li2021temporal}      & 69.64 $\pm$ 10.64 & 0.68 $\pm$ 0.13\\ 	
LMDA  \cite{miao2023lmda}           & 88.31 $\pm$ 4.66 & 0.88 $\pm$ 0.05 \\ 	
Proposed Method         & \underline{91.92} $\pm$ 4.62  & \underline{0.92} $\pm$ 0.05  \\ \hline
\end{tabular}
\label{tab:sota_hgd}
% \end{scriptsize}
\end{table}

\textcolor{black}{In addition to the speech imagery task, a study has been conducted to analyze the impact of the proposed method on motor movement tasks. Motor movement decoding has various medical and non-medical applications \cite{aliakbaryhosseinabadi2017classification}. It can be used for cursor control, robotic arm control, augmented virtual surgery, and stroke rehabilitation \cite{hooda2020fusion,xu2021decoding}. The accuracy and f-measure comparison of the proposed method with state-of-the-art methods is summarized in Table \ref{tab:sota_hgd}. The results of the proposed method are either better or more competitive than those of other state-of-the-art methods. It shows the robustness and generalizability of the proposed method in motor movement tasks. }

\begin{figure}[!t]
\centering
 \includegraphics[width=\linewidth]{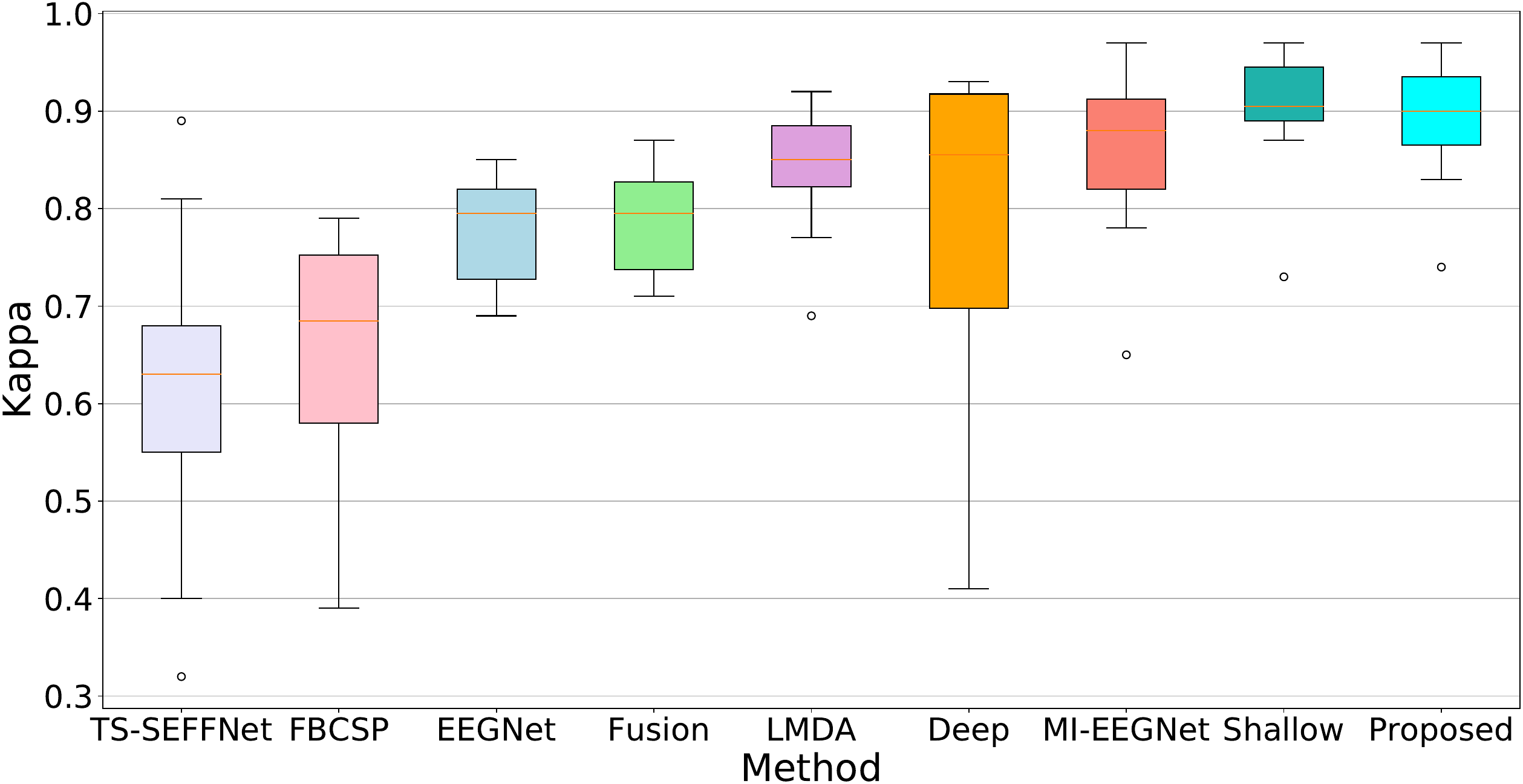}
  \caption{\textcolor{black}{Kappa measure comparison of proposed method and compared methods for motor movement dataset (MM-HGD dataset).} }
  \label{fig:kappa_hgd}
\end{figure}

\textcolor{black}{Kappa value comparison is shown in Figure \ref{fig:kappa_hgd}, and it indicates that proposed method have better class separability than all of the compared method. The only exception is the shallowconvnet method which has performed slightly better than the proposed method. However, the proposed method the difference of kappa value is very minute and it suggests that performance of the proposed method is competitive to shallowconvnet method. These results point out that the proposed method can be successfully utilized for motor movement tasks with high accuracy. }

\section{Discussion} 
\label{sec:discussion}
This section discusses the impact of various modules and hyper-parameters of the proposed method on the classification performance. Moreover, the activation maps generated using Grad-CAM method along with topographical plots are used to show the parts of the MI EEG signals considered significant by the proposed network. 
% The last subsection discusses about the performance of the proposed network on motor movement task.

\begin{figure}[!t]
\centering
 \includegraphics[width=\linewidth]{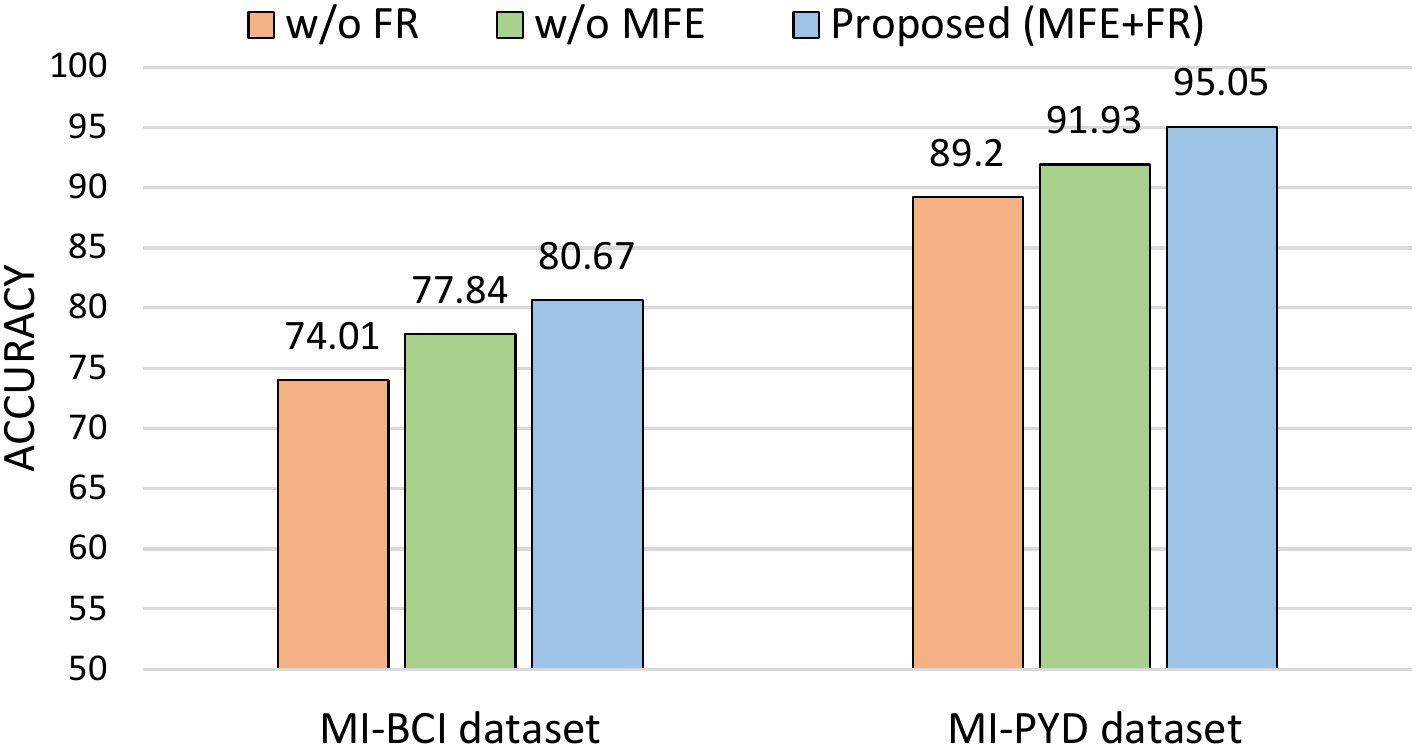}
  \caption{Ablation study to compute the impact of FR and MFE modules.} 
  \label{fig:ablation1}
\end{figure}

\subsection{Ablation study} 
In this section, a detailed ablation study is conducted to understand the effectiveness of each component of the proposed method. This ablation comprises an in-depth understanding of the impact of FR, MFE, TFS and CFS modules. Thereafter, the impact of the temporal scale factor ($\beta$) of TFS module and the channel split factor ($\gamma$) of CFS modules on the classification performance of the proposed method is discussed.

% \vspace{0.2cm}
\subsubsection{Impact of Feature Reweighting Module and Multi-Scale Feature Extraction Module}
In the proposed method, two important modules used for feature extraction are the multi-scale feature extraction (MFE) module and the feature reweighting (FR) module. In this ablation, the proposed method is compared with its two variants obtained by removing either MFE or FR module from the proposed model. The model without the MFE module is symbolized by w/o MFE, and the model without FR by w/o FR. FR + MFE symbolizes the proposed method. The performance of these models is depicted in Figure \ref{fig:ablation1}. 
It shows that the impact of FR is more than MFE in terms of MI-EEG classification accuracy in both datasets. It implies that the model without multi-scale feature extraction (w/o MFE) can extract more relevant features of the EEG signal and obtain better results. If the MFE is combined with the FR (i.e., the proposed method), it is better than the two variants for both datasets. It also shows the usefulness of the feature reweighting mechanism in getting better MI-EEG classification.

\begin{figure}[!t]
\centering
 \includegraphics[width=\linewidth]{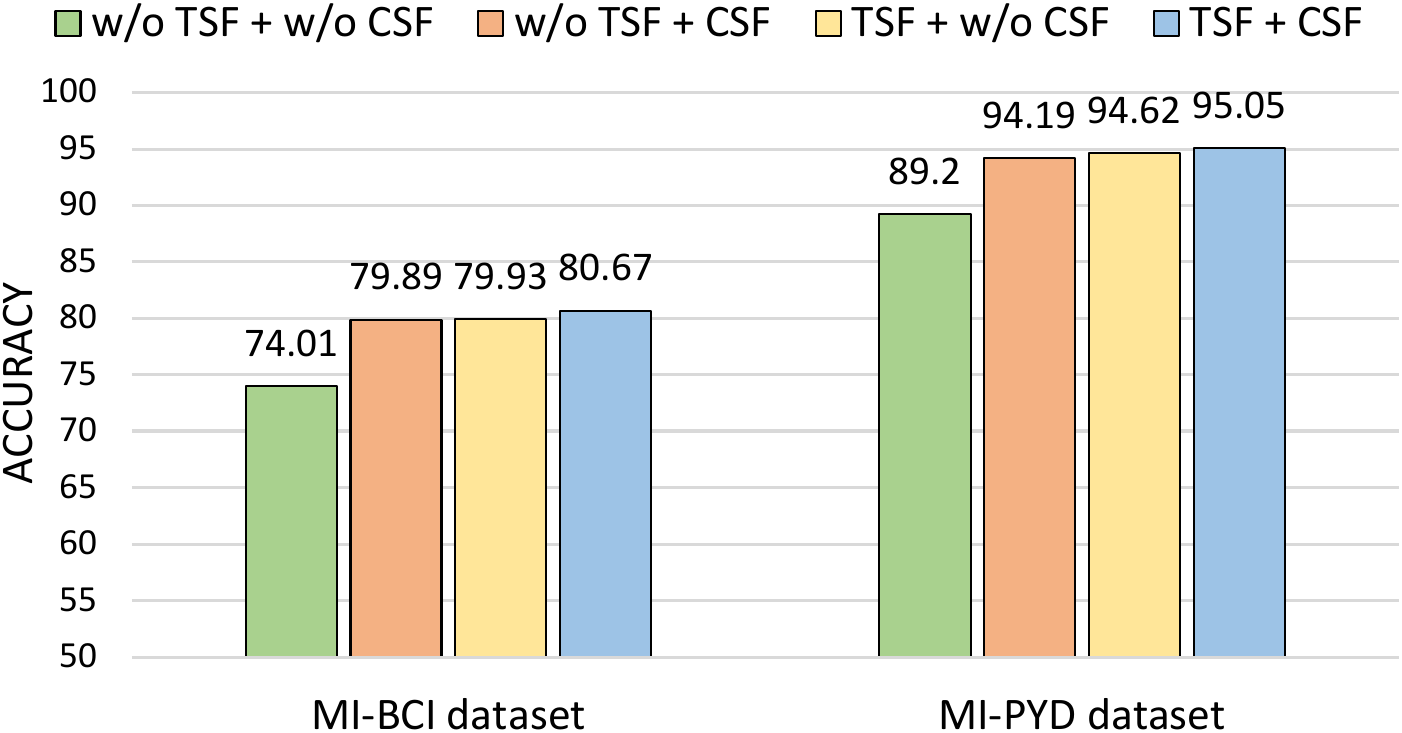}
  \caption{Ablation study to compute the impact of TFS and CFS submodules of Feature Reweighting Module.} 
  \label{fig:ablation2}
\end{figure}

\subsubsection{Impact of Temporal Feature Score Module and Channel Feature Score Module}

The feature reweighting mechanism is based on the score computed by TFS and CFS submodules. This ablation is conducted to explore the different combinations of the TFS and CFS and to observe their impact on MI-EEG classification. It results in four variants of the proposed method. The average classification performance of these four variants is summarized in Figure \ref{fig:ablation2}. Figure \ref{fig:ablation2} show that the variant that does not have TFS and CFS has the lowest classification performance for both datasets. Also, the TFS is found to be performing better than the CFS. Finally, adding TFS and CFS to the feature reweighting mechanism (i.e., the proposed method) helps to elevate the classification performance in both datasets.

\subsubsection{Impact of temporal scale and channel split factor}

\begin{figure}[!t]
\centering
 \includegraphics[width=\linewidth]{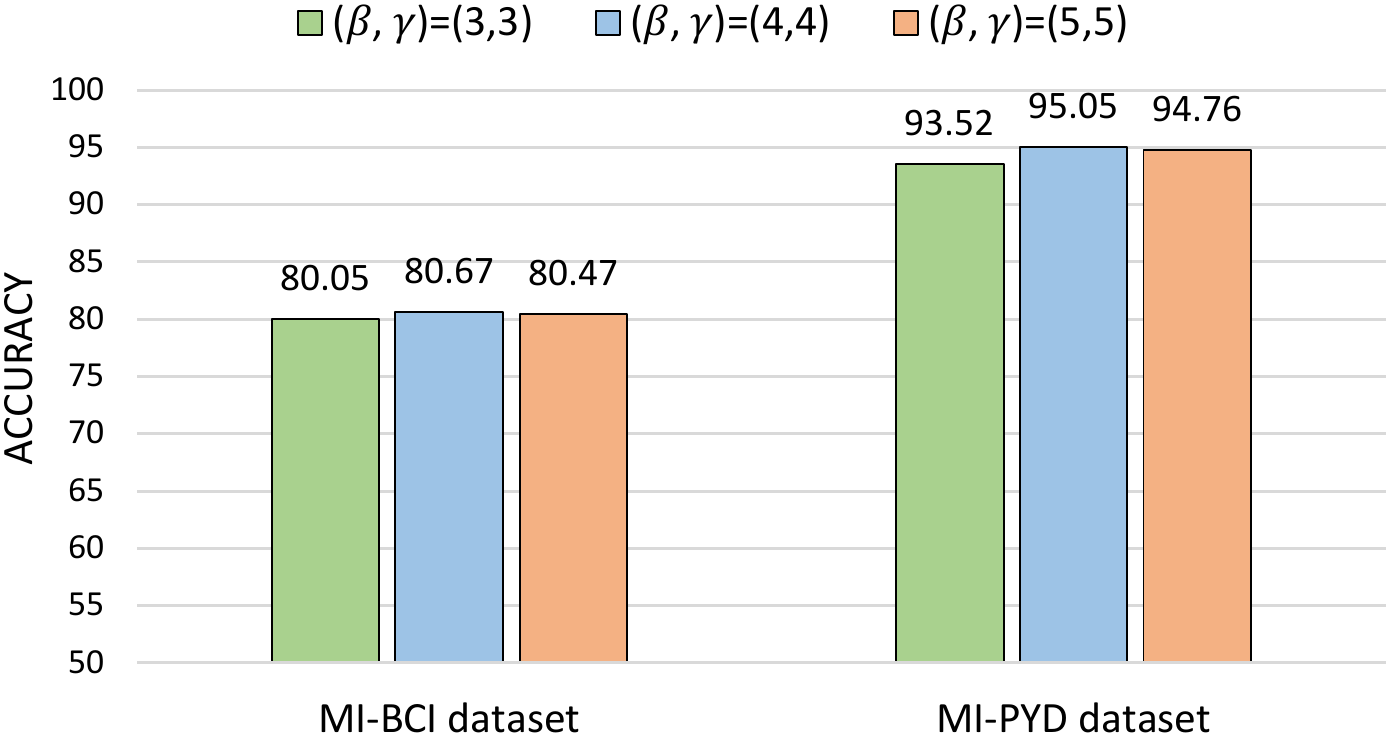}
  \caption{Ablation study on temporal scale and channel split factors ($\beta$ and $\gamma$) of FR module.} 
  \label{fig:ablation3}
\end{figure}

The proposed novel feature reweighting module relies on efficient temporal and channel score computation. The temporal and channel score computation module is designed over the temporal scale factor ($\beta$) and channel split factor ($\gamma$), respectively. Hence, the impact of these hyper-parameters in the network design is investigated in this ablation. The ablation results are shown in Figure \ref{fig:ablation3}. For ease of understanding, the same increment rate for both factors is followed in this study. From this ablation study, it can be observed that the optimal value for factors ($\beta$, $\gamma$) is ($4$, $4$) for both datasets (i.e., the value chosen for the proposed method). On further subject-wise analysis of the impact of factors, again, the factor ($4$, $4$) is preferred over the other factor values. Thus, it is concluded that the optimal value of the factors ($\beta$, $\gamma$) is ($4$,$4$) for this work.

\begin{figure*}[!t]
\centering
 \includegraphics[width=\textwidth]{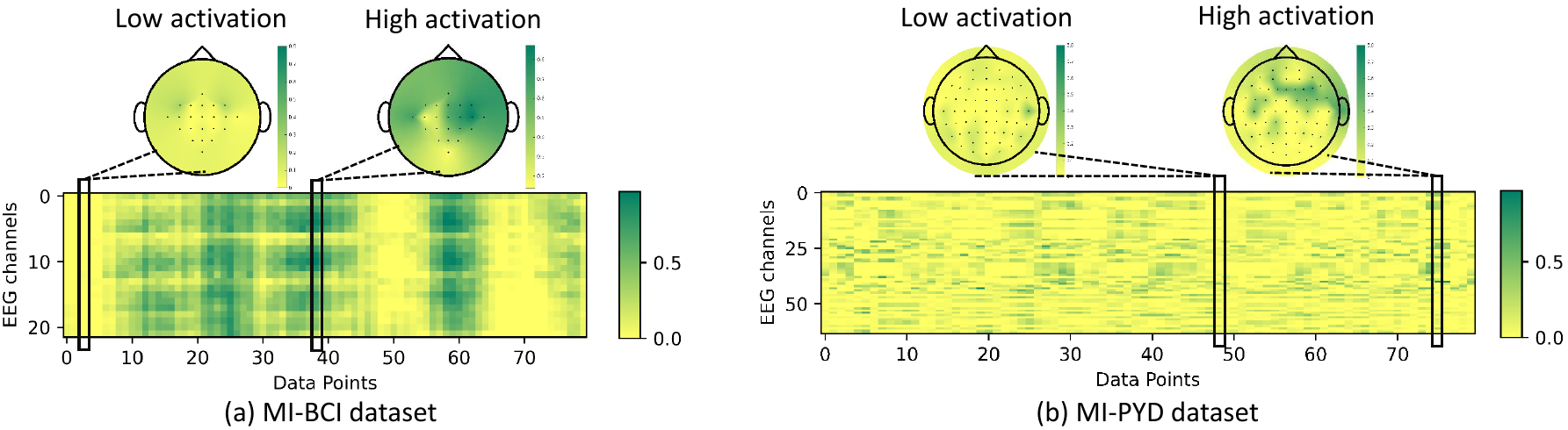}
  \caption{Grad-CAM activation maps with corresponding topographical plots of motor imagery task datasets. The brighter regions imply high attention, and the darker regions imply low attention to the input signal data points. \textcolor{black}{The black rectangles highlight the temporal datapoint, and the corresponding topographical map is shown above it.}}
  \label{fig:gradcam}
\end{figure*} 

% \vspace{-0.3cm}
\subsection{Grad-CAM visualization}
Grad-CAM is a popular method used to visualize the importance of the features extracted by a machine learning method~\cite{selvaraju2017grad}. The grad-CAM activation maps obtained from the proposed method are shown in Figure \ref{fig:gradcam}. 
Grad-CAM attention maps of both datasets show a repeated pattern of bright and dark regions. The protocol of the motor imagery task is the reason behind the observed pattern. In both datasets, the subjects were instructed to repetitively perform the particular motor imagery task until the trial ended. Therefore, the bright regions in the grad-CAM activation maps correspond to the time instances of brain activity, and the dark region corresponds to the time instances of brain inactivity. Additionally, topographical plots of low and high activation instances are shown above the grad-CAM activation maps in Figure \ref{fig:gradcam}. The high activation topographical plots of both datasets show increased activation in sensorimotor areas, whereas low activation plots show activity in other brain regions. It indicates that the proposed method gives high importance to the relevant features and low importance to irrelevant features of the MI-EEG signals.

\section{Conclusion}
\label{sec:conclusion}
This work presents a novel feature reweighting approach designed to suppress noisy and irrelevant features while emphasizing relevant ones in MI-EEG signals. The proposed method demonstrates superior performance compared to state-of-the-art approaches on two publicly available motor imagery classification datasets. It achieved classification accuracy improvements of 3.82\% on MI-BCI dataset and 9.34\% on MI-PYD dataset over CNN-based methods. Additionally, network variants incorporating the Temporal Feature Score (TFS) and Channel Feature Score (CFS) modules outperformed CNN-based methods, with accuracy gains of 3.08\% and 2.04\% on MI-BCI dataset, and 8.91\% and 8.48\% on MI-PYD dataset, respectively. These findings underscore the importance of reducing noise and irrelevant information across both temporal and channel dimensions in improving CNN-based classification. \textcolor{black}{Moreover, the proposed method showed superior performance in speech imagery tasks and competitive results in motor movement tasks, further highlighting its robustness and generalizability.} Future work could extend the method to additional EEG-based BCI paradigms for classification applications.

% The feature reweighting mechanism helps to provide high importance to the relevant temporal and channel features maps of CNN-based network. 

% The proposed method has shown significant improvement over the compared methods.
% by a significant margin compared to the baseline approach. 

% \bibliographystyle{ieeetr}
% \bibliography{paper.bib} 

\section*{Acknowledgements}
The first author gratefully acknowledges support from the Junior Research Fellowship awarded by the University Grants Commission (UGC), India (No. 3646/(NET-DEC2018)). The authors would also like to thank the anonymous reviewers and the editor for their valuable comments and suggestions.

 \bibliographystyle{elsarticle-num} 
 \bibliography{paper}





% \end{thebibliography}
\end{document}